\documentclass{article}

\usepackage{arxiv}

\usepackage[utf8]{inputenc} 
\usepackage[T1]{fontenc}    
\usepackage[hypertexnames=false]{hyperref}       
\usepackage{bookmark}
\usepackage{xurl}           
\usepackage{url}            
\usepackage{amsfonts}       
\usepackage{nicefrac}       
\usepackage{microtype}      
\usepackage{lipsum}
\usepackage{graphicx}
\usepackage{amsmath}
\usepackage[numbers]{natbib}

\usepackage{booktabs}             
\usepackage{multirow}             
\usepackage{makecell}             
\usepackage{tabularx}             

\usepackage{algorithm}
\usepackage{algorithmicx}
\usepackage{algpseudocode}
\usepackage{listings}

\makeatletter
\providecommand*{\toclevel@algorithm}{0}
\makeatother

\graphicspath{ {./images/} }

\title{VAMP-Net: An Interpretable Multi-Path  Network of Genomic Permutation-Invariant Set Attention and  Quality-Aware 1D-CNN for MTB Drug Resistance}

\author{
 Aicha Boutorh $\thanks{Corresponding author}$ \\
National School of Artificial Intelligence (ENSIA), \\ 
Sidi Abdellah Campus,  Algiers, Algeria\\
  \texttt{aicha.boutorh@ensia.edu.dz} \\
   \And
 Kamar Hibatallah Baghdadi \\
  National School of Artificial Intelligence (ENSIA), \\ 
  Sidi Abdellah Campus,  Algiers, Algeria\\
\texttt{kamar.baghdadi@ensia.edu.dz} \\
  \And
 Anais Daoud \\
  National School of Artificial Intelligence (ENSIA), \\ 
  Sidi Abdellah Campus,  Algiers, Algeria\\
  \texttt{anais.daoud@ensia.edu.dz} \\
}

\begin{document}
\maketitle
\begin{abstract}
Genomic prediction of drug resistance in Mycobacterium tuberculosis is often hindered by complex epistatic interactions and variable sequencing quality. We present the Interpretable Variant-Aware Multi-Path Network (VAMP-Net), a novel architecture addressing these challenges through a dual-pathway approach. Path-1 utilizes a Set Attention Transformer to model permutation-invariant variant sets and capture epistatic dependencies, while Path-2 employs a 1D-CNN to analyze VCF quality metrics for adaptive confidence scoring.Evaluated on four critical anti-TB drugs, VAMP-Net significantly outperforms baseline CNN and MLP models, achieving accuracies $>$ 95\% and AUCs $\sim$0.97 for Rifampicin and Rifabutin. Feature attribution analysis via Integrated Gradients successfully recovered canonical targets (\textit{rpoB, embB, katG}) and discovered high-impact novel loci. Functional enrichment confirmed these novel variants constitute non-random metabolic modules (p=0.00239) centered on cell-wall remodeling.Furthermore, systematic ablation of the Quality-Aware pathway demonstrates that the model performs a learned "integrated audit," prioritizing the Fraction of Supporting Reads and relative confidence over raw depth to mitigate technical noise. This dual-layer interpretability, bridging genomic pathogenicity with technical reliability, establishes a new paradigm for robust, auditable, and clinically actionable resistance prediction, positioning VAMP-Net as an important  tool for both diagnostic classification and mechanistic discovery in clinical genomics. 
\end{abstract}



\section{Introduction}
\label{intro}

The rapid growth of high-throughput sequencing has transformed genomics into a data-rich discipline, offering unprecedented opportunities to decode complex biological phenomena like drug resistance in \textit{Mycobacterium tuberculosis} (MTB) \citep{sharma2022tuberculosis}. For clinical applications, drug resistance prediction from whole-genome sequencing (WGS) must be both accurate and interpretable. However, modeling MTB resistance presents two primary challenges: (1) the genetic basis is governed by high-order epistatic interactions, requiring models that learn set-based relationships rather than simple sequential dependencies; and (2) clinical WGS data varies significantly in quality, making it crucial to distinguish true pathogenic variants from sequencing artifacts. Effectively balancing biological causality with data reliability is essential for developing robust predictive systems \citep{james2025whole, kim2022machine, cosgun2020exploring}. \\

Recent deep learning advances, particularly Convolutional Neural Networks (CNNs) and Transformers, offer promising avenues for capturing non-local relationships in genomic data \citep{perelygin2025deep, wu2023transformer}. While CNNs excel at local pattern recognition, they often require fixed-length inputs and struggle with the long-range epistatic interactions inherent in drug resistance. Conversely, while Transformers can model these dependencies, their application to unordered sets of genomic variants remains sparse. Furthermore, existing models typically treat critical variant calling (VCF) quality metrics as binary hard filters, discarding potentially valuable continuous information that could modulate the model's confidence in a variant's biological contribution \citep{eren2023improving}.\\

Despite recent advancements, existing models frequently treat genomic variants as rigid sequences and rely on static, "hard" VCF filters. This approach is limited to account for the fundamentally \textbf{permutation-invariant} nature of genomic variant sets and the continuous spectrum of sequencing quality metrics. Current methods either ignore technical noise entirely or rely on brittle, expert-defined thresholds that cannot capture the nuanced relationship between data confidence and predictive certainty.\\

To address these gaps, we propose \textbf{VAMP-Net}, an interpretable multi-path fusion architecture designed to bridge the gap between genomic deep learning and clinical microbiology. Our key contributions are summarized as follows:

\begin{itemize}

\item Expert-Informed Multi-Path Fusion: We introduce a dual-stream architecture that decouples biological signals from technical noise. By integrating a Set Attention Block (SAB) pathway with a quality-aware 1D-CNN, VAMP-Net mirrors the "clinical audit" workflow of expert microbiologists, allowing technical confidence to dynamically modulate biological predictions through a Gated Adaptive Fusion mechanism.

\item Permutation-Invariant Modeling via SAB: Unlike sequence-based Transformers, our model utilizes the \textbf{Set Attention Block (SAB)} to capture high-order epistatic interactions. This approach enforces a permutation-invariant inductive bias, ensuring the model respects the biological topology of the isolate as an unordered set of variants rather than an artificial sequence.

\item Clinically-Grounded Interpretability: We provide a validation framework that maps Path-1 attention weights directly to established resistance hotspots (e.g., \textit{katG}, \textit{rpoB}). By demonstrating that high-attention regions align with verified microbiology, we confirm that VAMP-Net’s decision-making is driven by biological causality rather than spurious statistical artifacts.
\end{itemize}

The rest of the paper is organized as follows. Section \ref{relatedwork} summarizes related work. Section \ref{method} details the proposed Multi-Path Network. Section \ref{results} presents the experimental results, interpretability analysis, and discussion. Section \ref{conclusion} summarizes our contributions and future perspectives.


\section{Related Work}
\label{relatedwork}

The evolution of drug resistance prediction in \textit{M. tuberculosis} (MTB) has progressed from simple association studies to complex deep learning frameworks. This section categorizes prior efforts and identifies the specific gaps addressed by the proposed Multi-Path architecture, as summarized in the comparative overview in Table \ref{tab:comparative_review}.

\subsection{Statistical and Classical Machine Learning}Genome-Wide Association Studies (GWAS) remain the foundation for identifying single nucleotide polymorphisms (SNPs) linked to drug resistance \citep{sharma2022tuberculosis, sun2021revisiting}. Traditional methods, however, struggle with high-dimensional data and non-linear epistatic interactions \citep{sigala2023machine}. To overcome these, penalized regression techniques such as LASSO and Ridge have been employed to identify significant loci \citep{an2020genome, maciukiewicz2018gwas}. Ensemble methods, including Random Forests (RF) and Gradient Boosting Machines (GBM), have demonstrated robustness to genomic noise \citep{nicholls2020reaching, ma2022sts}. Hybrid approaches have further enhanced performance by using RF for dimensionality reduction followed by clustering or Support Vector Machines (SVM) for classification \citep{silva2022machine, gaudillo2019machine}.  Prior studies in \citep{boutorh2016complex, boutorh2015classication, boutorh2014grammatical} focused on dimensionality reduction through hybrid frameworks based on association rule mining (ARM) and ANN to address the ``large $p$, small $n$'' challenge. More recently, \citep{deelder2019machine} and \citep{saliba2025enhanced} introduced flexible machine-learning and group-association frameworks to jointly model correlated resistance phenotypes, while systematic evaluations of ML strategies on mutation-based inputs have highlighted the utility of standardized feature sets \citep{paredes2025predicting}. Despite their utility, these models often rely on expert-defined feature engineering and treat sequencing quality metrics as binary filters rather than informative features.

\subsection{Deep Learning and Monolithic Architectures}Deep learning (DL) has significantly advanced post-GWAS prioritization by capturing non-linear interactions through architectures like DeepSEA and ExPecto \citep{zhou2015predicting, zhou2018deep}. Convolutional Neural Networks (CNNs) have become a standard for predicting phenotypic traits from SNP data due to their ability to model local patterns \citep{liu2019phenotype, romagnoni2019comparative}. Specifically for MTB, models like MD-CNN and CNN-GWAS have successfully integrated CNNs with genomic analysis to address missing heritability \citep{green2022convolutional, kwon2022genome}. Other frameworks have utilized monolithic deep structures, such as DeepAMR \citep{yang2019deepamr}, which employed denoising autoencoders to learn latent representations of co-occurrent resistance, and MD-WDNN \citep{chen2018deep}, which pioneered a wide-and-deep approach to share information across multiple drugs. Furthermore, 
 GenTB \citep{groschel2021gentb} and TB-DROP \citep{wang2024tb} provided fast, clinically-applicable resistance calls using multi-task and MLP-based representations. While successful, these models often rely on fixed-length mutation vectors that struggle with the ``long-tail'' of rare variants and show limitations to distinguish between biological signals and technical sequencing noise.

\subsection{Transformer-based and Graph Modeling}The emergence of Transformer architectures, such as DNABERT \citep{ji2021dnabert}, has enabled the processing of large DNA sequences through self-attention mechanisms. Lightweight models like LOGO combine self-attention and convolutional layers to interpret non-coding regions with high resolution \citep{yang2022integrating}. Modern sequence models like LLMTB \citep{testagrose2025leveraging} leverage BERT-based tokens to capture complex genomic patterns. This shift toward attention-driven models is reflected across clinical diagnostics, from cardiovascular disease detection \citep{noor2025novel} to oncology-driven feature extraction \citep{siddique2024optimizing}. Additionally, graph-based approaches have been introduced to model relational dependencies; for example, HGAT \citep{yang2021end} employs heterogeneous graph attention networks to model the topological relationships between isolates and mutations, while \citep{boutorh2022graph} utilized graph representation learning for drug repurposing.

\subsection{The VAMP-Net Position: Expert-Informed Decoupling}Despite the sophistication of current state-of-the-art (SOTA) models, a significant gap remains: existing architectures typically treat genomic data as a uniform black box, limited to isolate biological causality from technical artifacts. As illustrated in Table \ref{tab:comparative_review}, VAMP-Net addresses this critical limitation through an expert-informed multi-path architecture. Unlike the sequence-dependent logic of LLMTB or the fixed-graph constraints of HGAT, VAMP-Net utilizes a permutation-invariant Set Attention Transformer (Path-1) to capture non-linear epistatic interactions as unordered biological sets. Crucially, it introduces a dedicated Technical Auditor (Path-2) using a 1D-CNN to process quality-aware metrics, a feature absent in the monolithic designs of DeepAMR or DeepPTB. By integrating these streams via an Adaptive Fusion mechanism, VAMP-Net mirrors the clinical audit-trail logic used by microbiologists, providing a level of diagnostic reliability and auditable interpretability that current models struggle  to achieve.

\begin{table}[!htbp]
\centering
\footnotesize 
\setlength{\tabcolsep}{3pt} 
\caption{Comparison of architectural properties and training datasets across prior genomic modeling approaches versus VAMP-Net.}
\label{tab:comparative_review}
\begin{tabular}{@{}lllccl@{}}
\toprule
Method & Model Type & Variant Repr. & Quality Handling & Interpretability & Dataset Used \\
\midrule
WHO/GWAS & Statistical & Categorical SNPs & Hard Thresholds & High (Manual) & \makecell[l]{CRyPTIC \\ ($N\approx 10k$)} \\ \addlinespace
GenNet & CNN & Hierarchical Image & Binary Filter & Feature Map & \makecell[l]{Public Repos \\ ($N\approx 20k$)} \\ \addlinespace
DeepAMR & Autoencoder & Binary Matrix & None & Latent Space & \makecell[l]{Global Cohort \\ ($N\approx 16k$)} \\ \addlinespace
MD-WDNN & Wide \& Deep & Fixed-Length Vector & Static Input & Weights & \makecell[l]{PATRIC/ReSeqTB \\ ($N\approx 3.6k$)} \\ \addlinespace
HGAT & GNN & Heterogeneous Graph & Implicit & Edge Importance & \makecell[l]{Kaggle/AMK \\ ($N\approx 4.4k$)} \\ \addlinespace
DNABERT & Transformer & k-mer Sequence & N/A & Self-Attention & Ref. Genomes \\ \addlinespace
LLMTB & Transformer & BERT Tokens & Sequence-only & Attention Maps & \makecell[l]{CRyPTIC \\ ($N\approx 12k$)} \\ \addlinespace
TB-DROP & MLP/CNN & Mutation Matrix & Fixed Filters & Low & \makecell[l]{Curated Repos \\ ($N\approx 4k$)} \\ \midrule
\textbf{VAMP-Net} & \textbf{Multi-Path} & \textbf{Unordered Set} & \textbf{Adaptive Fusion} & \textbf{Gated Logic} & \textbf{\makecell[l]{CRyPTIC \\ ($N\approx 10k$)}} \\
\bottomrule
\end{tabular}
\end{table}


\section{Methodology}
\label{method}

Our approach introduces a Variant-Aware Multi-Path Network (\textit{VAMP-Net}) that synergistically couples a Set Transformer architecture with a quality-aware 1D-Convolutional Neural Network for genetic variant analysis. This network is designed as  a two-branch fusion model presented in Figure \ref{fig:model_architecture}. 


\subsection{VAMP-Net Overview:}

\textbf{A).Path-1:} The first branch, a Set Transformer, is specifically engineered to process the genetic variant sequence as an unordered set of tokens. This branch employs the Set Attention Block (SAB) \citep{lee2019set} to capture the complex non-sequential relationships between the variants, reflecting the inherent permutation-invariant nature of a set. \\

\textbf{B). Path-2:} The second branch of our framework is a quality-aware 1D-CNN. This path focuses on local sequential features. For each variant, it takes as input a corresponding quality score sequence.  The 1D-CNN is efficient in capturing the pattern in these local quality signals, thus offering the model critical information on the confidence and reliability of a variant call. 

\begin{figure}[H]
\centering
\includegraphics[width=1\linewidth]{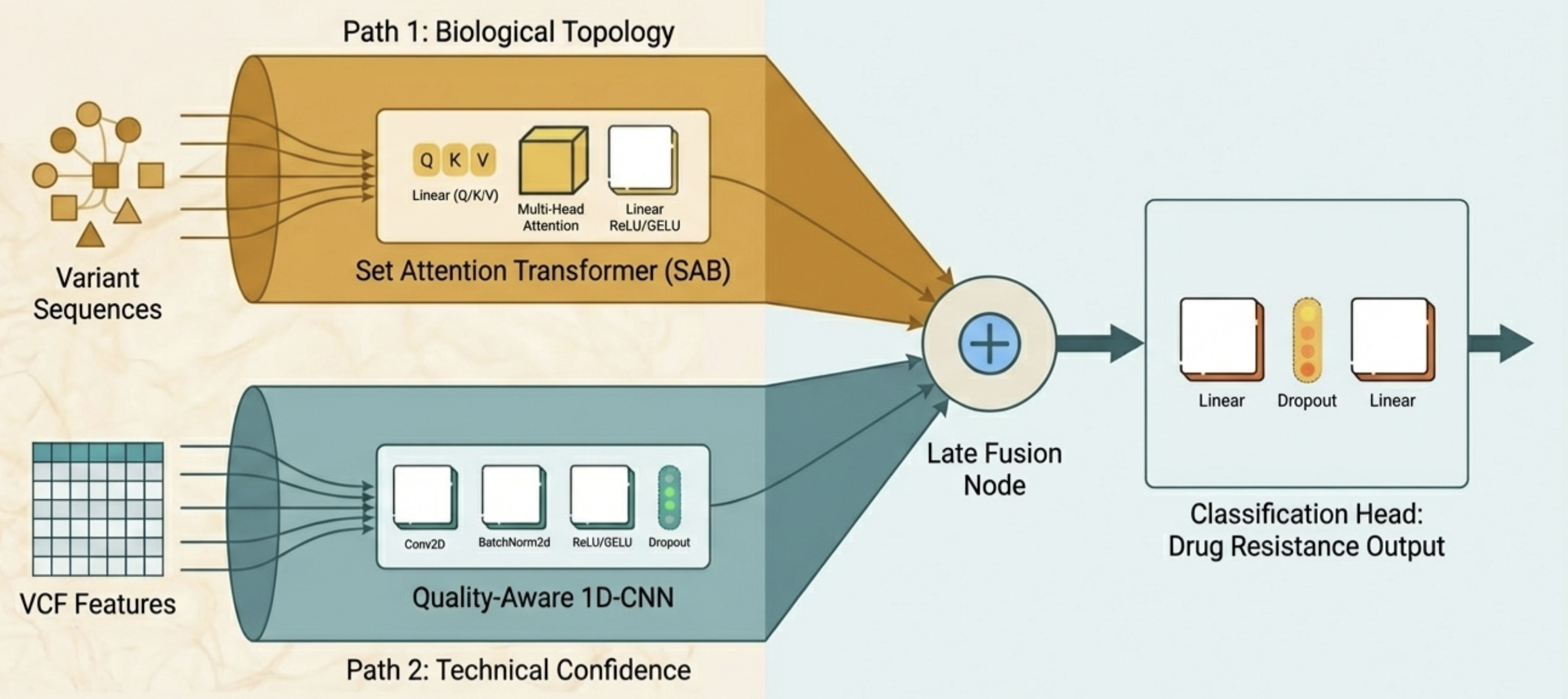}
\caption{Overview of the \textbf{Variant-Aware Multi-Path Network (VAMP-Net)} Architecture. The framework employs a dual-pathway design to disentangle biological signal from technical noise. \textbf{Path 1 (Biological Topology)} utilizes a Set Attention Transformer (SAB) to process genomic variant sequences as unordered sets; the multi-head attention mechanism captures high-order epistatic interactions between mutations while maintaining permutation invariance. Simultaneously, \textbf{Path 2 (Technical Confidence)} employs a Quality-Aware 1D-CNN to process multi-channel VCF features, extracting local patterns of sequencing reliability. The outputs of both paths are integrated at a \textbf{Late Fusion Node} via an adaptive gating mechanism, allowing technical confidence to modulate biological findings. The fused representation is finally processed by a \textbf{Classification Head} consisting of fully connected layers and dropout regularization to produce the final drug resistance prediction.}
\label{fig:model_architecture}
\end{figure}

\textbf{C). Fusion Module:} Finally, the outputs of the Set Transformer and the 1D-CNN branches are fused through a late-fusion mechanism, as pesented in the Figure \ref{fig:model_architecture}. The  representations from both branches are concatenated and passed to a final classification layer. This fusion allows the model to make predictions based on a rich, multi-faceted understanding of the data: the global, set-based relationships between variants from the SAB, and the local, sequential quality signals from the 1D-CNN. 
This multi-scale learning (local + global) provides both robustness (resistance to noisy data) and biological interpretability (attention highlighting key variants).\\

To formulate the problem addressed, a genomic sample is represented as a set of 
$N$ genetic variants: $  V= \{v_1, v_2, ..., v_N\} $  where each variant $v_i$ is described by  $(\textbf{g}_i, \textbf{f}_i)$,  with  $\textbf{g}_i$ representing a sequence of genetic alteration  and $\textbf{f}_i \in \mathbb{R}^d$ denoting the associated feature vector extracted from Variant Call Format (VCF) files. The prediction task is binary classification: $y = f_{\theta}(V)$, where $y \in \{0, 1\}$ indicates resistance or susceptibility to a given drug, and $f_{\theta}$ is a parameterized deep learning model. The VAMP-Net approach is presented in Algorithm \ref{alg:vampnet}.


\begin{algorithm}
\caption{VAMP-Net Multi-Path Inference and Adaptive Fusion, performed on an NVIDIA H100 GPU.}
\label{alg:vampnet}
\begin{algorithmic}[1]
\Require Set of Genomic Variants $\mathcal{V} = \{v_1, v_2, \dots, v_n\}$, VCF Quality Metrics $\mathbf{Q} \in \mathbb{R}^{n \times d}$
\Ensure Resistance Probability $\hat{y} \in [0, 1]$ \\

\Statex \textbf{// Path 1: Biological Signal Extraction}
\State $\mathbf{E} \gets \text{CompositionalTokenizer}(\mathcal{V})$ \Comment{Handle OOV via sub-tokenization}
\State $\mathbf{Z}_{bio} \gets \text{SetAttentionBlock}(\mathbf{E})$ \Comment{Capture permutation-invariant epistasis} \\

\Statex \textbf{// Path 2: Technical Context Assessment}
\State $\mathbf{Z}_{tech} \gets \text{1D-CNN}(\mathbf{Q})$ \Comment{Extract local patterns from VCF metadata} \\

\Statex \textbf{// Adaptive Fusion and Gating}
\State $\mathbf{g} \gets \sigma(\mathbf{W}_f [\mathbf{Z}_{bio} \parallel \mathbf{Z}_{tech}] + \mathbf{b}_f)$ \Comment{Compute adaptive gating vector}
\If{Mechanism is \textit{Suppression}}
    \State $\mathbf{Z}_{fused} \gets \mathbf{Z}_{bio} \odot \mathbf{g}$ \Comment{Down-weight low-confidence signals}
\ElsIf{Mechanism is \textit{Amplification}}
    \State $\mathbf{Z}_{fused} \gets \mathbf{Z}_{bio} + (\mathbf{Z}_{bio} \odot \mathbf{g})$ \Comment{Enhance high-confidence signals}
\EndIf \\

\Statex \textbf{// Final Prediction}
\State $\mathbf{h} \gets \text{ReLU}(\mathbf{W}_h \mathbf{Z}_{fused} + \mathbf{b}_h)$
\State $\hat{y} \gets \text{Sigmoid}(\mathbf{W}_{out} \mathbf{h} + b_{out})$
\State \Return $\hat{y}$
\end{algorithmic}
\end{algorithm}

\subsection{Expert-Informed Design: The Rationale for Functional Decoupling}

The choice of these two specific architectures  is a mathematical and biological necessity based on the different "shapes" of the genomic data.  Genomic data in a clinical setting is heterogeneous. It contains two entirely different types of information that should not be processed the same way. \\

\textbf{1.} \textit{Why Permutation-Invariant Set Attention? (The Biology)}:

\begin{itemize}

    \item The Nature of Mutations: Genomic variants in an isolate function as a biological set rather than a linear sequence. Whether a mutation in katG appears before or after a mutation in rpoB in the data file is irrelevant to the bacteria's resistance.

   \item Inductive Bias: Standard Transformers or RNNs assume a sequence (like a sentence). By using Set Attention (SAB) and removing positional encodings, we enforce a "set" inductive bias. This tells the model: "\textit{Focus on which mutations are present together (epistasis), not the order they were recorded.}"

   \item Capturing Epistasis: Attention mechanisms are mathematically superior at capturing non-linear interactions between distant variants (e.g., how a compensatory mutation in ahpC interacts with a primary mutation in katG).

\end{itemize}

\textbf{2.} \textit{Why 1D-CNN? (The Technical Auditor:}

\begin{itemize}

    \item The Nature of Sequencing Quality: VCF features (Depth, Fraction of Supporting Reads, Confidence scores) are local, quantitative signals.

    \item Pattern Extraction: 1D-CNNs are the gold standard for extracting features from local neighborhoods of multi-channel data. They are designed to detect "shapes" in the noise, such as a specific drop in read depth across a variant, that signify a low-quality call.

    \item Feature Differentiation: If we fed these technical metrics into the Transformer along with the variants, the high-magnitude "quality" numbers might overwhelm the sparse "biological" signals. The 1D-CNN acts as a pre-processor that compresses raw technical noise into a single "Confidence Vector."

\end{itemize}

\textbf{3.} \textit{Biomimetic Rationale: The Clinical Audit Workflow:}

Beyond mathematical necessity, the dual-pathway architecture of \textit{VAMP-Net} is explicitly designed to mirror the diagnostic reasoning of expert clinical microbiologists. In a manual diagnostic setting, an expert does not evaluate a mutation in isolation; instead, they engage in a "multi-path" heuristic:

\begin{itemize}

    \item \textbf{Biological Assessment (Path-1):} The expert identifies the presence of specific genetic markers (e.g., \textit{katG} or \textit{rpoB} mutations) and evaluates their known association with drug resistance.

    \item \textbf{Technical Auditing (Path-2):} Simultaneously, the expert scrutinizes the raw sequencing run's quality, verifying that the read depth and genotype confidence are sufficient to support a clinical call.
\end{itemize}

By formalizing these distinct tasks into specialized architectural branches, \textit{VAMP-Net} avoids the "black-box" pitfalls of monolithic models. This functional decoupling ensures that the final prediction is not merely a statistical correlation, but a clinical decision where the biological signal is rigorously audited by technical reliability, a process we term \textbf{Expert-Informed Fusion}.


\subsection{Dataset for MTB Drug Resistance:}
\label{datareepresentatiion}

The dataset used in this study comprises genetic variant sequences of Mycobacterium tuberculosis (MTB). Each sample is represented by a set of all identified genetic variants, resulting in sequences of variable length.


\subsubsection{Data Description }

Data for this study were obtained from  Comprehensive Resistance Prediction for Tuberculosis: an International Consortium (CRyPTIC) \citep{cryptic2022data}. This globally important resource consists of a robust set of 12,289 clinical isolates of Mycobacterium tuberculosis, collected from 23 countries. Importantly, the dataset provides matched whole-genome sequencing (WGS) information with quantitative Minimum Inhibitory Concentration (MIC) measurements for 13 essential antitubercular drugs, covering the entire treatment landscape: first-line agents (rifampicin, isoniazid, ethambutol), key second-line drugs (amikacin, kanamycin, rifabutin, levofloxacin, moxifloxacin, ethionamide), and newer/repurposed therapeutics (bedaquiline, clofazimine, delamanid, linezolid).\\

The genetic data is provided in variant call format (VCF) files with variants mapped against the \textit{M. tuberculosis} H37Rv reference genome (NC000962.3). Each VCF file contains several key features: 

\begin{itemize}
    
\item GT (genotype) indicates the called allele at each position; 
\item DP (depth) represents total read depth at the variant site; 
\item COV\_REF and COV\_ALT provide reference and alternate allele coverage respectively; 
\item DPF (depth fraction) indicates the proportion of reads supporting the variant; 
\item GT\_CONF (genotype confidence) gives the likelihood ratio for the called genotype; 
\item GT\_CONF\_PERCENTILE normalizes this confidence score; and 
\item FRS (a custom score) reflects variant quality. 
\end{itemize}

These features allow for detailed quality assessment of variant calls and enable filtering based on sequencing depth, allele balance, and call confidence.\\

The phenotypic data includes both continuous MIC values and binary resistance classifications (resistant/susceptible) based on epidemiological cutoff values, with 88\% of isolates having complete profiles across all 13 drugs. The dataset is enriched for drug resistance, with 55.4\% of isolates resistant to at least one drug. The complete dataset is publicly available at \url{ftp.ebi.ac.uk/pub/databases/cryptic/release_june2022/}.


\subsubsection{Data Representation: Genetic Variants as a Set of Tokens}
Transformer-based architectures have fundamentally transformed the processing of sequential data, exhibiting remarkable ability to capture long-range dependencies and contextual relationships. In the genomics domain, recent transformer models have achieved promising results; however, most existing approaches primarily rely on reference DNA sequences or \textit{K-mer} tokenization as inputs \citep{consens2025interpreting}. While such representations effectively capture local sequence patterns, they often treat variants as abstract tokens embedded within a continuous sequence. This abstraction can obscure the explicit biological relationships and combinatorial interactions among individual genetic alterations, relationships that are essential for understanding complex traits such as drug resistance. \\

To address this limitation, we propose a variant-centric representation that explicitly encodes both the molecular identity and genomic position of each variant as a single, biologically meaningful token. In this framework, each token directly represents a specific reference-to-alternate allele substitution, enabling the model to learn variant-level interactions rather than merely sequence-level dependencies. This transition from sequence-level attention to variant-level interaction attention provides a more interpretable and biologically grounded foundation for modeling genotype–phenotype associations. Specifically, we employ an attention mechanism that operates on a $ChromPos\_Ref>Alt$ representation, designed to capture biologically meaningful variant–variant interactions. In this representation, each genetic variant $\textbf{a}_k$ is uniquely identified by its chromosome, precise genomic position, reference allele, and alternate allele (e.g., \textit{chr1:12345:A$>$G}). This compact and canonical identifier serves as a semantically rich token, allowing each sample’s genomic profile $\textbf{g}_i$ to be expressed as a discrete set of variant tokens \textit{T}: $  \textbf{g}_i= \{a_1, a_2, ..., a_T\} $. Such a representation enables the transformer to reason directly over meaningful biological entities, facilitating the discovery of high-order, non-linear relationships among variants that underlie phenotypic outcomes such as drug resistance.\\

\begin{figure}[H]
\centering
\includegraphics[width=1\linewidth]{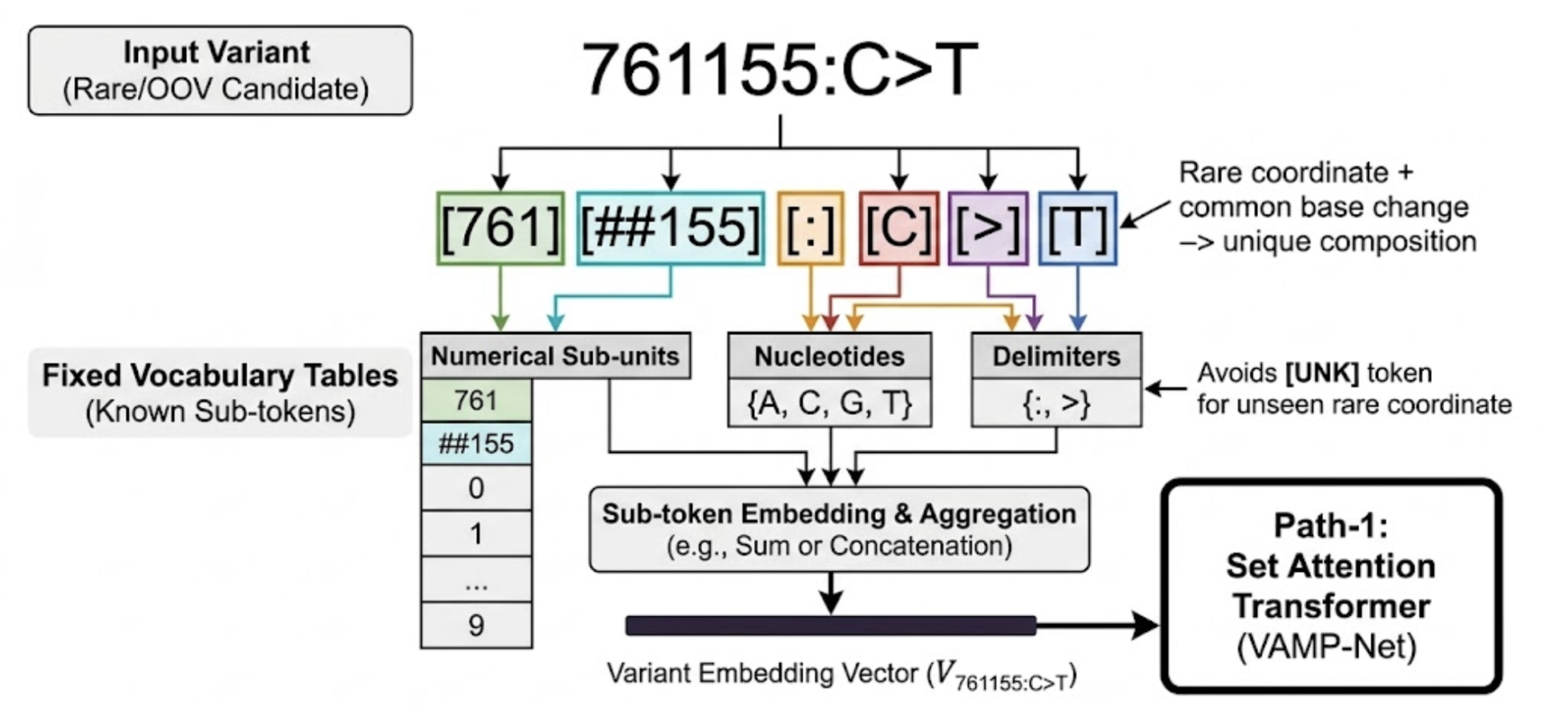}
\caption{  WordPiece-based tokenization strategy for genomic variant encoding. This figure details the mechanism used to mitigate Out-of-Vocabulary (OOV) errors during the processing of non-canonical mutations. An input variant (e.g., $761155:C>T$) is processed via WordPiece-based tokenization, which decomposes the symbolic string into fundamental sub-units: numerical subunits, nucleotides, and delimiters. By leveraging Fixed Vocabulary Tables that contain these universal genomic components, VAMP-Net can construct unique embedding vectors for previously unseen rare coordinates. This approach avoids the information loss associated with traditional $[UNK]$ (unknown) tokens. The resulting \textbf{Sub-token Embeddings} are aggregated to produce a single, high-dimensional \textbf{Variant Embedding Vector} ($V_{761155:C>T}$), which is then passed to the Path-1 Set Attention Transformer. This method ensures that the model can learn from the "long tail" of genomic data while maintaining a compact and efficient vocabulary.}
\label{fig:token}
\end{figure}

To mitigate out-of-vocabulary (OOV) effects, we employ a data-driven tokenization scheme in which a BERT WordPiece tokenizer is trained on the complete set of unique variant strings extracted from the dataset As illustrated in Figure \ref{fig:token}. The tokenizer operates without lowercasing and learns a compact vocabulary composed of genomic primitives and frequently occurring subpatterns, including numeric positions, canonical nucleotides (A, C, G, T), and structural delimiters (e.g., $>$ and $:$). At inference time, each variant string is decomposed into valid subword units using a greedy longest-match strategy, ensuring that even rare or previously unseen variant compositions within the same representational schema can be expressed through combinations of known tokens. This approach substantially reduces reliance on the $[UNK]$ token and preserves the compositional structure of genomic variants. The resulting sub-token embeddings are subsequently aggregated to form a unified high-dimensional representation of each variant, which is then provided as input to the downstream Path-1 Set Attention Transformer. By aligning the tokenizer with the generative structure of the data, this method enables robust modeling of the long-tail distribution of genomic variants while maintaining vocabulary efficiency.

\subsection{Model Architecture: Variant-Aware Multi-Path Network (VAMP-Net)}
\label{modelarchite}

Variant-Aware Multi-Path Network (\textit{VAMP-Net}), a dual-path deep learning network specifically designed to address the limitations of conventional single-stream models in genomic data analysis. \textit{VAMP-Net} is architected to process two distinct sources of information essential for robust variant-based prediction ( Figure \ref{fig:model_architecture}):
(1) the symbolic identity and biological context of each genetic variant and
(2) the quality metrics of quantitative sequencing  that reflect data reliability. \\

The rationale for this dual-path design is twofold: first, to leverage the inherent set-based nature of a sample's variant, where order is non-critical, by using a modern attention mechanism; and second, to capture local patterns and dependencies within the associated quality signals using convolutional processing.\\

The model is structured as follows:

\begin{itemize}
    \item Path-1 – Symbolic and Topological Stream:
Processes the set of genomic variants using a symbolic tokenization scheme at variant-level, allowing the model to learn relationships regardless of input order.

\item Path-2 – Feature and Sequential Stream:
Processes a structured sequence of quantitative and contextual features (e.g., read depth,  genotype quality) using one-dimensional convolutional layers to extract local quality patterns.

\end{itemize}

The outputs from both pathways are subsequently fused to form an integrated latent representation, which is passed to a final prediction head. This fusion allows \textit{VAMP-Net} to jointly reason over global, biologically meaningful variant interactions and local, quality-aware signals, thereby offering a comprehensive and interpretable foundation for genomic variant modeling.


\subsubsection{Path-1:  Set Attention Transformer Block (SAB) on Genetic Variants }
\label{path1}

\begin{figure}[H]
\centering
\includegraphics[width=1\linewidth]{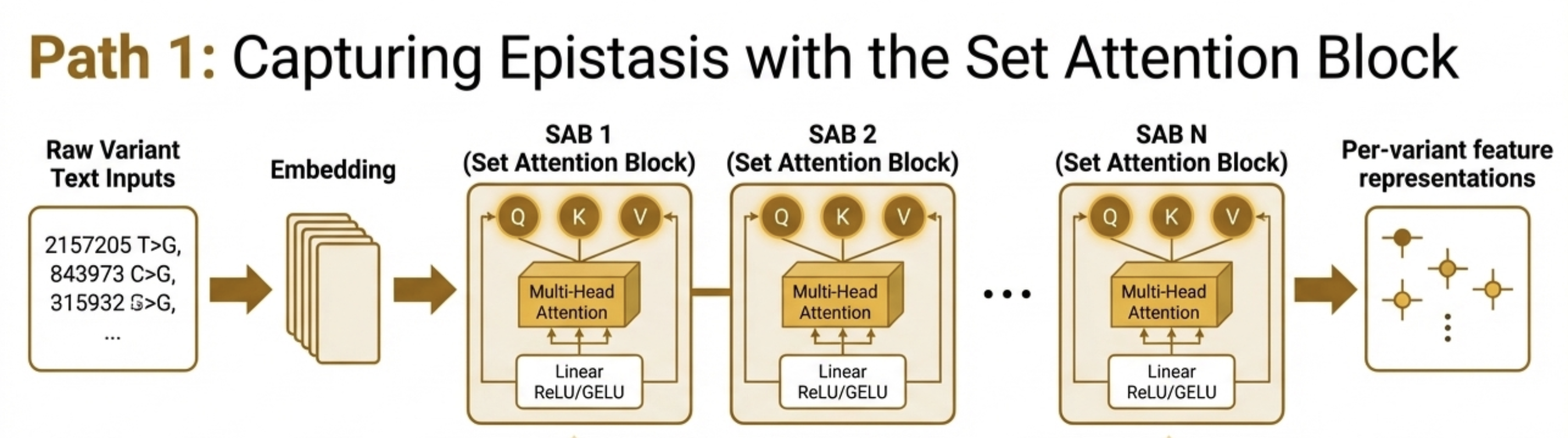}
\caption{\textbf{Path-1 Architecture: Set Attention Transformer Block (SAB) for Symbolic Variant Processing}. This pathway facilitates the extraction of high-order epistatic features from raw genomic data. Raw variant text inputs (formatted as $Position\_Ref>Alt$) are first mapped into a continuous vector space via a \textbf{Compositional Embedding} layer. These embeddings are processed through a series of $N$ stacked \textbf{Set Attention Blocks (SAB)}. Unlike standard Transformers, the SAB utilizes self-attention mechanisms without positional encodings, ensuring the model remains \textbf{permutation-invariant}. Within each SAB, Multi-Head Attention (MHA) computes Query (Q), Key (K), and Value (V) interactions to model the non-linear dependencies between distant mutations. The final output is a set of \textbf{per-variant feature representations} that encode the biological topology of the isolate, specifically capturing the synergistic effects of multiple mutations on the drug resistance phenotype.}
\label{fig:SABarchi}
\end{figure}

We address the modeling of genetic variant sequences for downstream tasks of drug resistance prediction. Each sample $\mathcal{X}_i$ consists of a variable-sized set $\mathcal{X}_i$ = $\{x_1, x_2, ..., x_T\}$ where each element $x_k$ represent a specific genetic variant. Since $\mathcal{X}_i$ is an unordered set, the architecture must satisfy \textbf{permutation invariance}, formulated in Equation \ref{eq:permutation}:

\begin{equation}
\label{eq:permutation}
f(\pi\mathcal{X}_i) = f(\mathcal{X}_i), \quad \forall \pi \in S_{n},
\end{equation}

where $\pi$ denotes a permutation operator on the sequence elements.
The first branch of our model is a Set Transformer, a deep learning architecture specifically designed to operate on sets presented in Figure \ref{fig:SABarchi}. Its core component, the Set Attention Block (SAB), replaces the standard transformer's multi-head attention mechanism. Unlike standard transformers which rely on positional encodings and often causal masks for sequential data, the SAB is inherently permutation-invariant. This means the model's output is independent of the order of the input tokens. For our genetic variant sequences,  where the order is arbitrary, the attention mechanism’s ability to model interactions among all elements in a set without
relying on their positions is a critical advantage. This enables the model to learn a unified representation of the set, a property that is essential for our application.\\ 

The SAB pathway models variant-variant interactions while maintaining permutation invariance. For an input sequence represented by a matrix $\textbf{X} \in \mathbb{R}^{n \times d_{model}}$, where $n$ is the sequence length and $d_{model}$ is the embedding dimension, the SAB computes attention scores between all pairs of tokens in the input sequence, followed by a weighted sum of the values. Mathematically (Equation \ref{eq:attention}), for an input set of tokens:

\begin{equation}
\label{eq:attention}
\text{Attention}(\textbf{Q}, \textbf{K}, \textbf{V}) = \text{softmax} \left( \frac{\textbf{Q}\textbf{K}^{\top}}{\sqrt{d_k}} \right) \textbf{V}
\end{equation}

where $d_k$ is the dimension of the keys. Unlike standard transformers, \textbf{no causal mask} is applied  to preserve permutation invariance. The output of SAB is aggregated using a pooling operator $\rho$ as formulated in Equation \ref{eq:sab}:
\begin{equation}
\label{eq:sab}
\textbf{z}_{\text{SAB}} = \rho(\text{SAB}(\textbf{X}))
\end{equation} \\

We apply the Set Attention Block (SAB) in two distinct configurations, which were investigated to assess the impact of explicit padding utilization: \\

 \textbf{I - Unmasked Set Attention:} \\

The first contribution adapts the Set Transformer framework \citep{lee2019set} of the unmasked SAB  for modeling genetic variant sequences represented as unordered tokens. Unlike standard transformers for sequential data, no attention masks —neither causal  masks is applied, ensuring each variant can attend to every other variant without positional or causal constraints, nor padding masks is applied. This model serves as a theoretically pure baseline for evaluating the efficacy of a purely permutation-invariant approach. The input to the model is a batch of padded sequences, where each variant is represented by its token embedding. Within the SAB, the self-attention mechanism computes attention scores for every token pair, including those involving the meaningless padding tokens. The model learns to encode the relationships between all tokens, with no explicit instruction to ignore the padded ones. This approach tests the model's capacity to learn a meaningful representation of the data despite the presence of noisy, non-informative tokens. The training objective is to perform a specific task of Drug Resistance, based on their set of variants. The hyperparameters for this model, including the number of SAB layers, hidden dimensions, and learning rate, are provided in the Experimental Section \ref{sec:model_experiments}. \\

 \textbf{II- Padding-Masked  Set Attention :}\\
 
 The second contribution is the introduction of a padding mask to the attention mechanism, a crucial modification that   enables efficient batching and improves computational performance. The padding mask is a binary matrix that masks out attention to the padded tokens. It is applied by adding a large negative value to the attention scores of padded tokens prior to the softmax operation.\\

To handle sequences of varying lengths, we introduce in Equation \ref{eq:PadingMask} a \textbf{padding mask} $\textbf{M} \in \{0, -\infty\}^{N \times N}$, where:

\begin{equation}
\label{eq:PadingMask}
    M_{ij} =
    \begin{cases}
        0, & \text{if both } i, j \text{ are valid variants}, \\
        -\infty, & \text{if } j \text{ is padding}.
    \end{cases}
\end{equation}

The attention computation becomes as defined in Equation \ref{eq:Attentionmask}:

\begin{equation}
\label{eq:Attentionmask}
    \text{softmax} \left( \frac{\textbf{QK}^\top + \textbf{M}}{\sqrt{d_k}} \right) \textbf{V}
\end{equation}
which masks out contributions from padded positions but preserves invariance over valid variants. Since $M$ only removes invalid tokens (not reordering them), the equivariance property still holds over the subset of valid elements (Equation \ref{eq:EquivProp}):

\begin{equation}
\label{eq:EquivProp}
    \text{SAB}_{\text{masked}}(\textbf{P}\textbf{X}) = \textbf{P} \text{SAB}_{\text{masked}}(\textbf{X}),
\end{equation}

$\mathbf{P}$ a permutation matrix that reorders the tokens in $\mathbf{X}$. Provided $M$ is permuted identically as formulated in Equation \ref{eq:PermuteMatrix}. $\mathbf{M}'$ represents the permuted version of the padding mask $\mathbf{M}$ that must be used when the input tokens $\mathbf{X}$ are permuted by the matrix $\mathbf{P}$:

\begin{equation}
\label{eq:PermuteMatrix}
    M' = PMP^{\top}.
\end{equation}

The application of the padding mask $\mathbf{M}$ provides crucial theoretical and practical advantages for processing variable-length variant sets: 

\begin{itemize}
    \item Elimination of Irrelevant Signal: It ensures irrelevant (padded) tokens do not contribute to attention score computation,  thereby preventing the model from dedicating capacity to learning representations for non-existent variants.

\item Mitigation of Attention Dilution: By zeroing out the influence of padding, the mask prevents artificial self-attention dilution within smaller variant sets. This maintains the true relative importance of valid tokens, particularly when a large portion of the input is padding.

\item Preservation of Set Invariance: Since the mask is content-based (identifying the padding symbol) and not positionally fixed, it does not interfere with the core permutation invariance of the Set Attention Block, provided the mask is permuted identically with the input tokens.

\end{itemize}

\subsubsection{Path-2 : Quality-Aware 1D-CNN}
\label{Path2}

\begin{figure}[H]
\centering
\includegraphics[width=1\linewidth]{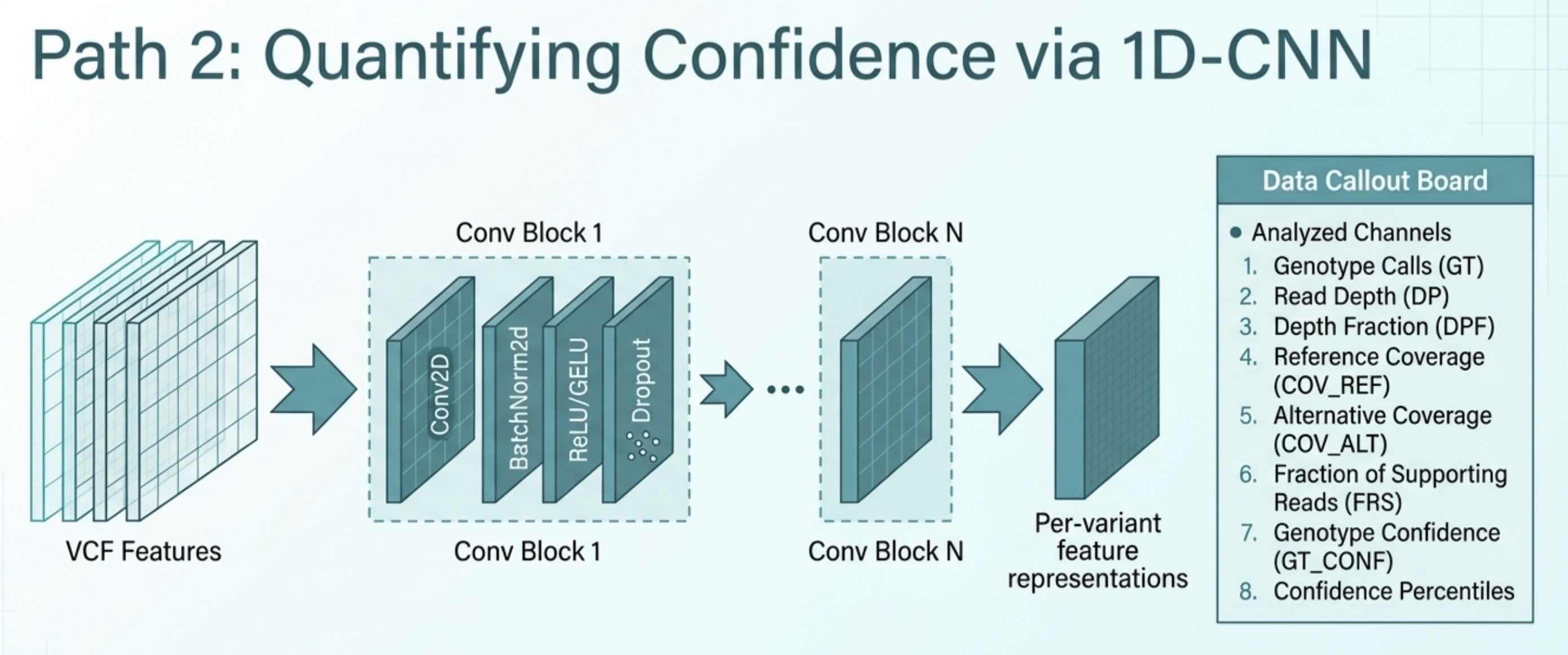}
\caption{\textbf{Path-2 Architecture: Quality-Aware 1D Convolutional Neural Network (1D-CNN) for VCF Feature Processing}. This pathway is dedicated to the quantitative assessment of sequencing reliability. The input consists of a multi-channel tensor of VCF-derived features, including \textbf{Genotype Calls (GT), Read Depth (DP), and Fraction of Supporting Reads (FRS)}, among others listed in the Data Callout Board. These features are processed through a series of $N$ stacked \textbf{Convolutional Blocks}, each containing a \textbf{1D-convolutional} layer for local pattern extraction. 
Unlike traditional binary filters, this 1D-CNN learns a continuous, \textbf{high-dimensional representation of technical confidence }for each variant. These per-variant feature representations are then utilized by the fusion module to adaptively weight the biological signals from Path-1, ensuring that resistance predictions are grounded in high-quality sequencing data.}
\label{fig:1D-CNNarchi}
\end{figure}

A second branch of our model consists of a 1D-CNN that processes the quality scores associated with each variant (Figure \ref{fig:1D-CNNarchi}). These scores, which are typically sequential and local, provide crucial information on the confidence of each variant call. A 1D-CNN is an ideal choice for this task, as it excels at capturing local patterns, such as the relationship between a variant's read depth (DP) and its genotype confidence (GT CONF). The 1D-CNN branch processes the variant feature matrix $\textbf{F} \in \mathbb{R}^{N \times d}$.\\

While the core genetic information (CHROM, POS, REF, ALT) is fundamental, the true richness of VCF files for deep learning lies within their FORMAT fields. These fields provide critical quantitative and qualitative information about the
confidence and context of each variant call, moving beyond a simple binary presence or absence of a mutation. This depth of information, often underutilized in conventional genomic deep learning approaches, represents a powerful, biologically relevant multi-channel input for sophisticated models. Using these detailed metrics, a deep learning model can gain a more nuanced understanding of the reliability and characteristics of each detected variant, which is paramount for accurate genotype-phenotype association studies, especially when dealing with potentially noisy or low-coverage sequencing data from real-world clinical samples.\\ 

The proposed framework specifically harnesses seven VCF FORMAT fields as distinct input channels for the  Convolutional Neural Network (1D-CNN). When extended to multi-channel inputs, a 1D-CNN can simultaneously process multiple features per genomic position, with each VCF field representing a distinct channel. This allows CNN to learn intricate local patterns and dependencies across these different quality and genotype metrics for each variant, rather than relying solely on the raw genotype information. 

\subsubsection{Multi-Path Fusion Mechanism}
\label{Fusion Mechanism}

The final stage of the \textit{VAMP-Net} architecture involves the formal integration of the learned representations from the two specialized paths: the biological topology manifold and the technical quality manifold.

\paragraph{- Formal Integration Framework:}
Let $\mathbf{z}_{\text{SAB}} \in \mathbb{R}^d$ be the permutation-invariant set representation from Path-1, and $\mathbf{z}_{\text{CNN}} \in \mathbb{R}^k$ be the quality-aware feature vector from Path-2. To ensure that technical confidence can modulate biological signal, we define a \textbf{Gated Fusion Module} that maps technical noise patterns to a modulation intensity vector $\mathbf{g} \in \mathbb{R}^d$ via a sigmoid-activated transformation:

\begin{equation}
\mathbf{g} = \sigma(\mathbf{W}g . \mathbf{z}_{\text{CNN}} + \mathbf{b}_g)
\end{equation}

where $\mathbf{W}_g$ and $\mathbf{b}_g$ are learnable parameters. This gating vector acts as a "Technical Auditor," determining the trustworthiness of each dimension in the biological feature space.

\paragraph{- Fusion Modulation Strategies:}
To determine the optimal strategy for combining these disparate signals, we investigate three distinct modulation mechanisms that resolve conflicts between biological data and technical noise:

\begin{enumerate}

\item 

\textbf{Suppression (De-emphasis of Noise):}

Inspired by the "forget gate" in Long Short-Term Memory (LSTM) networks, this mechanism performs element-wise multiplication to scale the biological signal:

\begin{equation}
\label{eq:SupFusion}
\mathbf{z}_{\text{fused}} = \mathbf{g} \odot \mathbf{z}_{\text{SAB}}
\end{equation}

As $\mathbf{g} \in [0, 1]$, this formulation performs a dynamic, data-driven feature selection. If Path-2 identifies poor sequencing quality (e.g., low depth), $\mathbf{g} \to 0$, effectively suppressing the influence of that variant to prevent the propagation of sequencing artifacts.

\item 

\textbf{Amplification (Emphasis on Signal):}
To test if high-quality data provides a critical signal gain, we introduce a baseline-preserving mechanism:

\begin{equation}
\label{eq:AmpFusion}
\mathbf{z}_{\text{fused}} = (1 + \mathbf{g}) \odot \mathbf{z}_{\text{SAB}}
\end{equation}

This ensures that the symbolic representation $\mathbf{z}_{\text{SAB}}$ is always preserved, while high-confidence variants receive a gain of up to $200\%$. This prioritizes robustly supported mutations without risking the "zeroing out" of potentially relevant signals.

\item 

\textbf{Adaptive Amplification and Suppression (Proposed Best):}

Our final strategy synthesizes both approaches into a bipolar scaling factor that provides the model with dynamic, context-aware control:

\begin{equation}
\label{eq:scaled_fusion}
\mathbf{z}_{\text{fused}} = (2\mathbf{g} - 1) \odot \mathbf{z}_{\text{SAB}}
\end{equation}

By mapping the scaling factor to the interval $[-1, 1]$, the model can continuously transition between a \textbf{Suppression Zone} (noise filtering) and an \textbf{Amplification Zone} (signal gain). This allows the model to not only ignore noise but to actively penalize uncertain variants by reversing their signal polarity, mirroring the rigorous audit-trail logic used by clinical microbiologists.
\end{enumerate}

\paragraph{Final Classification}
The resulting vector $\mathbf{z}_{\text{fused}}$ is passed through a global average pooling layer and a Softmax-activated dense layer to produce the final resistance probability $P(y=1)$:

\begin{equation}
P(y=1) = \text{Softmax}(\mathbf{W}f . \mathbf{z}_{\text{fused}} + \mathbf{b}_f)
\end{equation}

\paragraph{- Functional Interpretation of Adaptive Fusion:}

The bipolar scaling mechanism in VAMP-Net is designed to resolve conflicts between biological signals and technical noise. This mathematical operation corresponds to two distinct operational modes:

\begin{itemize}

\item \textbf{Adaptive Suppression:} If Path-2 detects poor sequencing quality (e.g., low depth or high genotype noise), the gating factor $\mathbf{g}$ approaches $0$, and the scaling factor becomes negative. This effectively filters the variant out, preventing the model from assigning resistance based on a potential sequencing artifact.

\item \textbf{Adaptive Amplification:} Conversely, when Path-2 confirms high technical confidence for a non-canonical mutation, the scaling factor approaches $1$. This amplifies the signal, allowing the model to prioritize novel, high-confidence epistatic interactions that might be missed by static filters.

\end{itemize}

This design ensures that biological causality is always contextualized by data reliability, mirroring the audit-trail logic used by clinical microbiologists during manual isolate review. 



\subsection{Interpretability}
\label{Interpretability}

Following the development of the Variant-Aware Multi-Path Network ($\text{\textit{VAMP-Net}}$), we conducted a critical interpretability analysis to identify genomic variants and sequencing features most significantly associated with \textit{RIF} and \textit{RFB} resistance in Mycobacterium tuberculosis. Our approach is uniquely tailored to the dual-pathway model architecture, employing methods that respect the inherent structure of each stream .

\subsubsection {Dual-Path Interpretability Framework}

We analyzed both pathways of the $\text{\textit{VAMP-Net}}$ architecture separately to isolate the contribution of symbolic variant identity from that of quantitative quality metrics: \\

\textbf{Path-1 (Set Transformer Analysis):} For the $\text{SAB}$ pathway, which processes the symbolically tokenized and permutation-invariant set of variants ($\mathbf{z}_{\text{SAB}}$), we applied Integrated Gradients (\textit{IG}). This method calculated attribution scores by integrating gradients from baseline input to actual variant embeddings, measuring how each genetic variant contributed to the final resistance prediction irrespective of its ordering in the input set. \\

\textbf{Path-2 (CNN Analysis):} 
For the convolutional pathway processing the ordered $\text{VCF}$ feature vector ($\mathbf{z}_{\text{CNN}}$), we employed gradient-based saliency analysis in conjunction with systematic feature ablation. The gradient analysis quantified the relative importance of feature channels (e.g., $\text{DP}$, $\text{GT}$, $\text{FRS}$) across the sequential input. Subsequently, systematic feature ablation confirmed the findings by quantifying the drop in predictive performance when specific VCF quality metrics were masked or perturbed. This framework reliably quantifies both the positional importance of variants along the input sequence and the absolute contribution of specific VCF quality metrics.


\subsubsection{Variant Importance Quantification}

For Path-1, which utilizes the Set Attention Transformer, we quantified the contribution of each individual variant by applying Integrated Gradients (IG)  with respect to the final classification layer output. We first extracted the variant-level embeddings from the trained symbolic tokenizer. The attribution scores were then calculated by integrating the gradients along a linear path over $N=20$ integration steps, ensuring robust and stable gradient estimates. Finally, the calculated attribution scores were aggregated across all instances in the test cohort to obtain the mean importance measure for each unique \text{ChromPos\_Ref$>$Alt} variant. This yields an interpretable ranking of variants based on their model-predicted association with drug resistance.


\subsubsection{Feature Ablation Studies}

To robustly quantify the influence of sequencing quality and context features (processed by Path-2), we conducted systematic feature ablation on individual VCF channels. This approach, designed to isolate the causal contribution of each metric, involved methodically setting the values of a single VCF feature channel to a neutral baseline (zero) across all test samples . We then rigorously measured the resulting change in the model’s prediction confidence and overall performance (e.g., AUC or $F_1$ score).\\

This systematic process allowed us to identify which VCF quality metrics were most critical for accurate resistance classification and, conversely, which features introduced minimal signal. The ablation analysis specifically covered the following eight VCF feature channels, reflecting key data quality and genotype metrics:
Genotype Calls ($\text{GT}$), Read Depth ($\text{DP}$), 
Depth Fraction ($\text{DPF}$), Reference Coverage ($\text{COV\_REF}$), Alternative Coverage, ($\text{COV\_ALT}$), 
Fraction of Supporting Reads ($\text{FRS}$), 
Genotype Confidence ($\text{GT\_CONF}$),
Confidence Percentiles ($\text{GT\_CONF\_PERCENTILE}$).\\

The resulting performance decrement for each ablated feature serves as its direct measure of importance within the $\text{VAMP-Net}$ architecture.


\subsubsection{Attention-Based Variant Interaction Analysis}

To move beyond individual variant importance and uncover complex epistatic relationships between resistance-associated variants, we leveraged the intrinsic relational mechanism of the Set Attention Block ($\text{SAB}$) layer from Path-1. The $\text{SAB}$'s self-attention matrix inherently encodes the model's learned pairwise dependency structure between every variant in the input set.\\

For each test sample, we systematically extracted the attention weights from the first $\text{SAB}$ layer and averaged them across all attention heads to compute the raw variant relationships. The interaction strength between any two variants was thus quantified as the bidirectional attention weight between their respective tokens, which was then aggregated across the entire test cohort for statistical reliability.\\

This aggregated matrix was subsequently used to construct a variant interaction network. Network construction was restricted to variants exhibiting sufficient frequency and strong, recurring interaction occurrences to ensure statistical reliability. Finally, we applied modularity optimization algorithms to the network to perform community detection, thereby identifying functional modules (groups of variants) that likely operate synergistically to confer drug resistance. The highly connected nodes identified in this process were designated as hub variants, representing key drivers of the overall epistatic effect.


\section{Experimentation and Discussion}
\label{results}

This section presents the empirical validation of the Variant-Aware Multi-Path Network ($\text{\textit{\textbf{VAMP-Net}}}$) and its core components, followed by a detailed discussion of the biological insights derived from our interpretability framework. We first detail the  preparation of the dataset and the specific encoding utilized for the $\text{ChromPos\_Ref$>$Alt}$ variant tokens. Subsequent subsections systematically evaluate three critical aspects of the design of $\text{\textit{VAMP-Net}}$: the selection of the optimal fusion mechanism, a comprehensive performance comparison, and a deep-dive interpretation of the predictions of the final model, including variant importance and epistatic interaction networks. The goal of this analysis is not only to demonstrate superior classification performance but to validate the translational utility of our dual-path approach for genetic resistance prediction.


\subsection{Data Preprocessing}
\label{subsec:data_preprocessing}

Prior to use, the data is subjected to multiple preparation procedures. After downloading the \texttt{CRyPTIC\_reuse\_table\_20240917.csv} file from the official compendium website, we  focus on four drugs: \textit{Rifampicin (RIF)}, which is primarily used to treat tuberculosis by inhibiting RNA polymerase;  \textit{Rifabutin (RFB)}, a related rifamycin derivative with similar therapeutic applications; \textit{Isoniazid (INH)}, a prodrug that targets the synthesis of mycolic acids, essential components of the mycobacterial cell wall; and \textit{Ethambutol (EMB)}, which acts by inhibiting the enzyme arabinosyltransferase, thereby disrupting the assembly of the mycobacterial cell wal. \\

First, we filter the dataset to include only high and medium quality samples, based on the \texttt{\{drug\}\_PHENOTYPE\_QUALITY} field. For each selected sample, we download its corresponding VCF file, which contains variant information. Then we remove columns with missing data and retain only entries marked as \texttt{PASS} in the \texttt{FILTER} field, indicating reliable variant calls.\\

To reduce dimensionality and reliability, we focus on positions where the ALT allele differs from the REF allele, discarding entries such as \texttt{0/0} and \texttt{0|0}, which indicate homozygous reference (no variant), as well as \texttt{./.} and \texttt{.|.}, which denote missing or uncalled genotypes, it should be noted that the slash (/) unphased genotype, while the pipe ($|$) indicates a phased genotype. Variants are represented as \texttt{{pos}\_{ref}$>${alt}} for all alleles separated by commas; we refer to this representation as the \textit{Path-1 Input}, after this filtering, we end up with a total of \textbf{\textit{417,635 unique variants}}. \\

We also retain additional descriptive features for each variant: \texttt{GT}, \texttt{DP}, \texttt{DPF}, \texttt{COV}, \texttt{FRS}, \texttt{GT\_CONF}, and \texttt{GT\_CONF\_PERCENTILE}. Numeric features are scaled using min–max normalization, while categorical features are encoded. The \texttt{FRS} field, representing the fraction of reads supporting the genotype call (\#reads supporting GT / total reads), ranges between 0 and 1; higher FRS values indicate stronger confidence in the genotype, whereas missing or low values typically correspond to low coverage or filtered calls, and hence we chose to replace its missing values with zeros. For each sample, these features are organised into an array of shape (number of variants × 8 features × 1), referred to as the \textit{Path-2 Input}, which will be used in a 1D-CNN model. Arrays are padded as needed to ensure uniform length across samples. \\

It should be noted that the resulting samples are not aligned with each other. For example, sample one may have \texttt{20\_A>G} at index 0, while sample two has \texttt{13\_C>T} at the same index. This indicates that the variants are not ordered strictly by their genomic positions in the input vectors. Originally, the variants were sorted by position in the VCF file, but after filtering and preprocessing, some variants shifted in the vectors. This positional inconsistency across samples motivates the use of a \emph{Permutation-Invariant Block (SAB)} in the model architecture.

\begin{figure}[H] 
\centering
\includegraphics[width=1\linewidth]{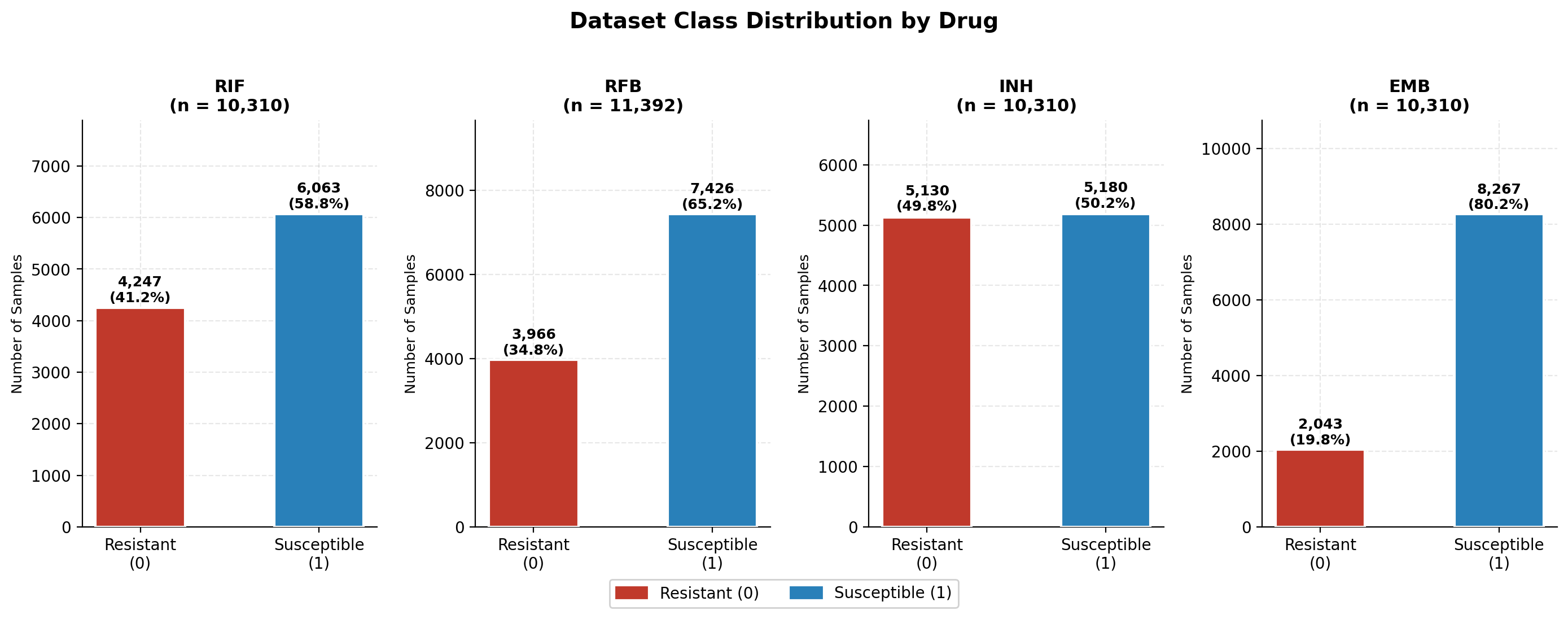}
\caption{Dataset Class Distribution across the selected anti-TB drugs. This figure illustrates the sample size ($n$) and the distribution between resistant (red) and susceptible (blue) classes for the four targeted drugs.}
\label{fig:drug_distribution}
\end{figure}

Figure \ref{fig:drug_distribution} illustrates the sample counts and class distributions for each drug. We observe a class imbalance of approximately 17.6\% between susceptible and resistant samples for RIF and 30.4\% for RFB, while EMB exhibits the most significant disparity at 60.4\%. To mitigate the effects of these imbalances during training, a weighted cross-entropy loss is employed \\

Initially, we utilize the \textit{RIF} dataset to evaluate various architectural design choices and establish an optimal modeling pipeline. This finalized approach is subsequently applied to \textit{RFB}, \textit{INH}, and \textit{EMB} to validate its generalizability across different drug resistance profiles.  The train/validation/test split was  (80/10/10), clarifying that this was performed at the isolate level to prevent data leakage. We also specify that stratification was applied based on the phenotypic resistance labels to ensure balanced representation across all sets.


\subsection{Experimental Results}

This section details a multi-stage evaluation of the VAMP-Net  to quantify its predictive performance and justify its architectural design. We first establish a performance baseline by evaluating various feature encoding strategies and fusion strategies.  A rigorous ablation study follows, isolating the specific contributions of the VCF-aware Path-2 branch to the model's decision-making process. Finally, we report the performance of the best-performing VAMP-Net configuration across four primary anti-TB drugs (RIF, RFB, INH, and EMB), evaluating its robustness, calibration, and clinical utility on unseen  isolates.

\subsubsection{Evaluation of Feature Encoding Strategies}
\label{encoding}

Following the preprocessing pipeline described in Section~\ref{subsec:data_preprocessing}, genomic variants are represented as $Pos\_REF$>$ALT$. We conducted a comparative evaluation between two distinct representational paradigms: (1) a Static Encoding approach, where each unique variant is mapped to a discrete identifier, and (2) a Contextualized Tokenization approach utilizing a BERT-based tokenizer trained on the corpus of unique variants to project genomic features into a continuous vector space.\\

\begin{figure}[H]
\centering
\begin{minipage}[b]{0.32\textwidth}
    \centering
    \includegraphics[width=\textwidth]{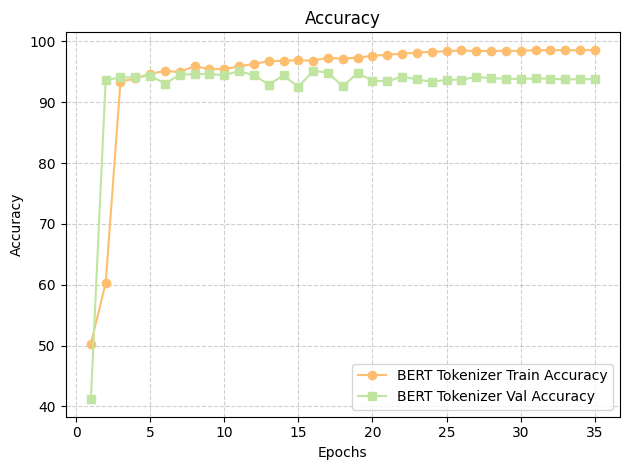}
\end{minipage}
\hfill
\begin{minipage}[b]{0.32\textwidth}
    \centering
    \includegraphics[width=\textwidth]{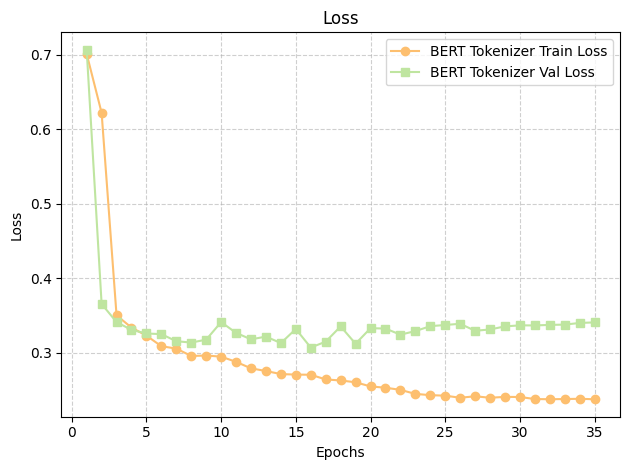}
\end{minipage}
\hfill
\begin{minipage}[b]{0.32\textwidth}
    \centering
    \includegraphics[width=\textwidth]{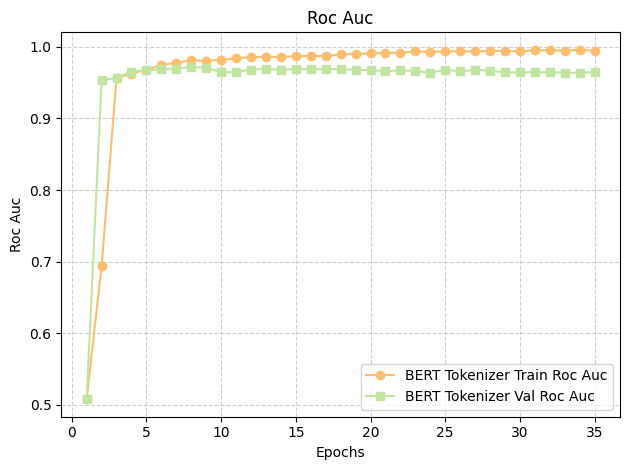}
\end{minipage}

\vspace{1em}

\begin{minipage}[b]{0.32\textwidth}
    \centering
    \includegraphics[width=\textwidth]{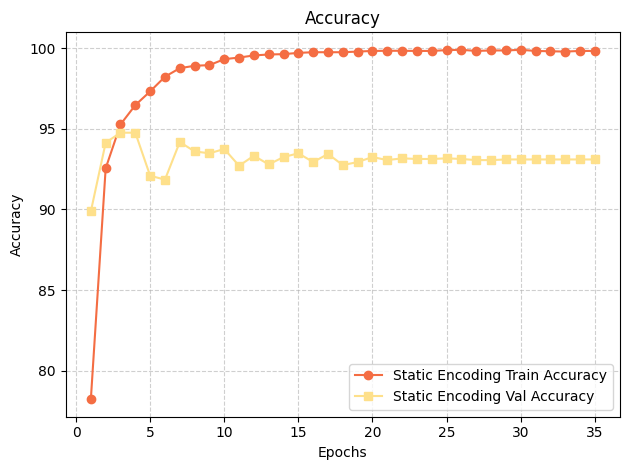}
\end{minipage}
\hfill
\begin{minipage}[b]{0.32\textwidth}
    \centering
    \includegraphics[width=\textwidth]{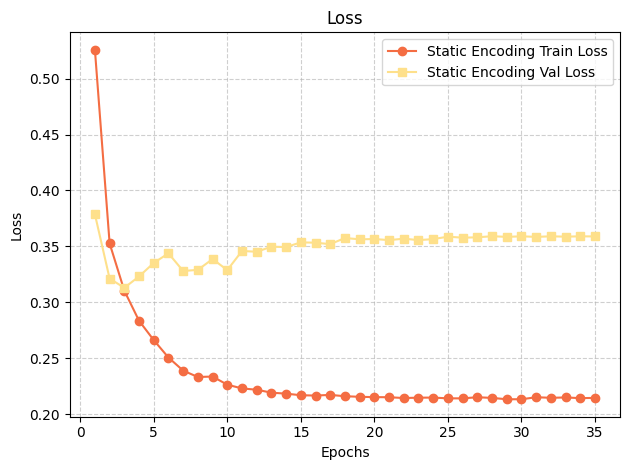}
\end{minipage}
\hfill
\begin{minipage}[b]{0.32\textwidth}
    \centering
    \includegraphics[width=\textwidth]{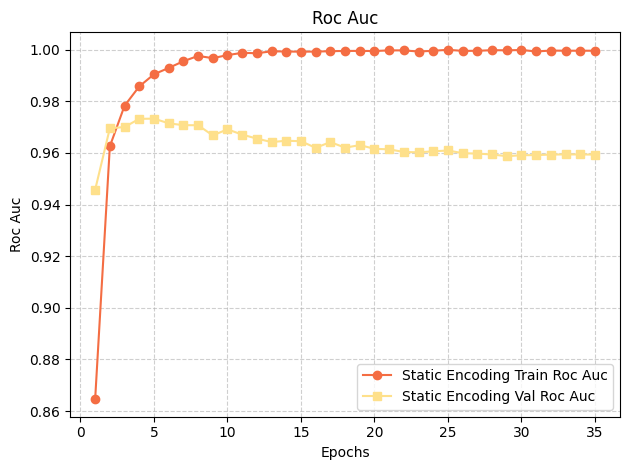}
\end{minipage}

\caption{Training and validation dynamics across tokenization strategies. Comparative analysis of BERT Tokenizer (top row) and Static Encoding (bottom row) across Accuracy, Loss, and ROC-AUC metrics. While both methods reach high training performance, the BERT-based approach demonstrates superior stability and generalization, as evidenced by the significantly narrower gap between training and validation curves in the Loss and ROC-AUC plots compared to the overfitting observed in Static Encoding.}
\label{fig:static_vs_bert}
\end{figure}

Empirical results indicate that the Static Encoding method suffered from premature convergence and poor generalization. Early stopping was triggered at epoch~3, with the learning curves exhibiting a significant $\sim$6.73\% generalization gap between training and validation accuracy. Conversely, the BERT-based representation demonstrated superior stability and robust feature extraction. While some degree of overfitting remained present—noted by a 4.83\% accuracy disparity—the model exhibited a more sustained learning trajectory, converging at epoch~16 with a smoother loss profile. As illustrated in Figure~\ref{fig:static_vs_bert}, the BERT-based encoding facilitated more reliable optimization; consequently, it was selected as the foundational encoding strategy for all subsequent experiments.

\subsubsection{Evaluation of Multimodal Late Fusion Strategies }

\begin{figure}[H]
\centering
\includegraphics[width=1\textwidth]{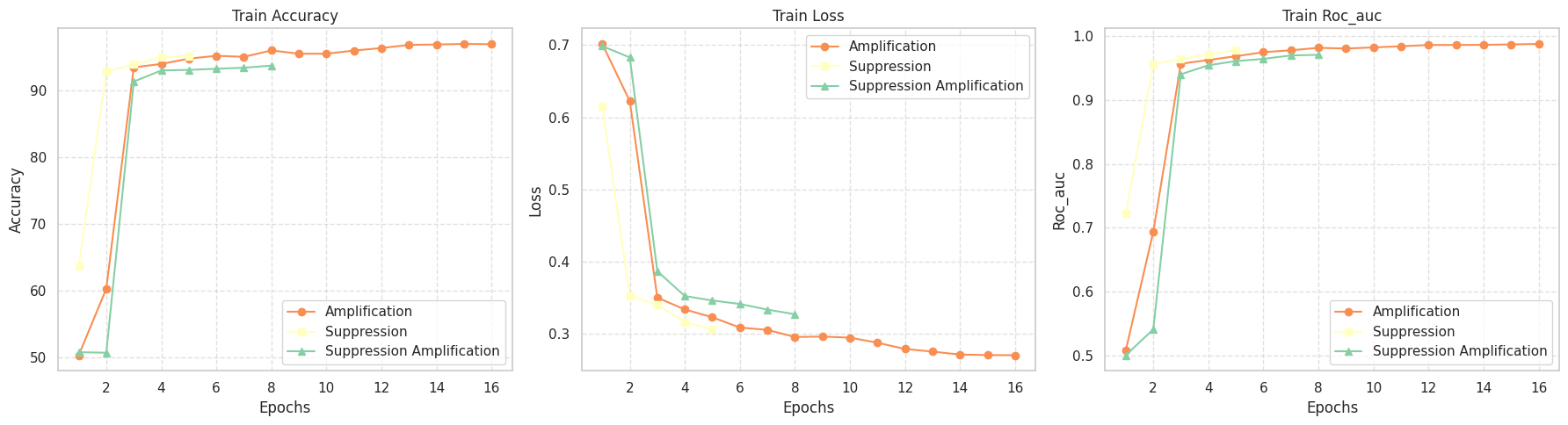}
\caption{Impact of Gated Fusion strategies on model convergence during training. The trajectories compare three fusion variants: Amplification (orange), Suppression (yellow), and the combined Suppression-Amplification mechanism (green). While all methods converge to high ROC-AUC values ($>0.95$), the Amplification strategy exhibits the most stable optimization profile over 16 epochs, whereas the combined Suppression-Amplification approach shows greater variance, suggesting a trade-off between strict noise filtering and signal preservation.}
\label{fig:fusion_curves}
\end{figure}

To identify the optimal integration strategy between Path-1 and Path-2, we evaluated three distinct fusion paradigms: Suppression, Amplification, and a hybrid Suppression-Amplification mechanism. The optimization trajectories are illustrated in Figure~\ref{fig:fusion_curves}, while a comprehensive performance summary across the test set is provided in Table~\ref{tab:fusion_table}.\\

\begin{table}[!htbp]
\begin{center}
\caption{Comparative performance metrics across Gated Adaptive Fusion mechanisms. The table evaluates three functional modes of the adaptive fusion head: Suppression, Amplification, and a combined Suppression-Amplification strategy.}
\label{tab:fusion_table}  
\begin{tabular}{@{}lccccc@{}}
\toprule
Fusion Type & Accuracy & Precision & Recall & F1 & AUC \\
\midrule
Suppression \& Amplification & 0.9267 & 0.9322 & 0.9439 & 0.9381 & 0.9572 \\
\textbf{Amplification} & \textbf{0.9523} & \textbf{0.9510} & \textbf{0.9598} & \textbf{0.9554} & 0.9690 \\
Suppression & 0.9438 & 0.9460 & 0.9591 & 0.9525 & \textbf{0.9710} \\
\bottomrule
\end{tabular}
\end{center}
\end{table}

While all three mechanisms demonstrated competitive performance with negligible variance in validation and testing metrics, the Amplification Fusion consistently yielded the highest predictive accuracy and exhibited superior generalization. Consequently, this strategy was adopted for the final model architecture, hereafter referred to as Model~A.


\subsubsection{Model Validation and Ablation Study }
\label{sec:model_experiments}

Building upon the selection of Model A as the architectural baseline, we conducted an extensive ablation study to evaluate the robustness of our padding/masking strategies, the efficacy of data augmentation, and the specific contributions of the dual-path integration. To systematically dissect the influence of each component, we evaluated six model configurations (Models A--F), as detailed below. The architectural variants were designed to isolate the effects of Attention Masking, Data Augmentation, and the Path-2 feature stream.

\begin{enumerate}
  \item \textbf{Model-A}: SAB  \underline{with} Attention Masks for Padding, \underline{without} Augmentation.  
  \item \textbf{Model-B}: SAB  \underline{without} Attention Masks for Padding, \underline{with} Augmentation.
  \item \textbf{Model-C}: SAB  \underline{without} Attention Masks for Padding, \underline{without} Augmentation.
  \item \textbf{Model-D}: SAB  \underline{without} Attention Masks for Padding, \underline{with} Augmentation.
  \item \textbf{Model-E}: Model-A settings, \underline{without} the inclusion of Path-2.  
  \item \textbf{Model-F}: Model-B settings, \underline{without} the inclusion of Path-2.  
\end{enumerate}

Optimal hyperparameters were identified via Bayesian Optimization (Tables \ref{table:hyperparams_pa}, \ref{table:hyperparams_pbch}), prioritizing a robust balance between the Area Under the Receiver Operating Characteristic (ROC AUC) curve and Balanced Accuracy on the validation set. This approach ensures that the selected configurations prioritize generalization over mere training set convergence.

\begin{table}[!htbp]
\begin{center} 
\caption{Comprehensive hyperparameter configurations for VAMP-Net. The table summarizes the optimized structural for Models A through F applied on RIF drug and model A apllied on RFB, INH and EMB drugs. This mapping includes the specific configurations for \textit{Path-1 Transformer (SAB).}}
\label{table:hyperparams_pa}
   
\begin{tabular}{@{}lccc@{}}
\toprule
Model & Emb Dim & Hidden Dim & Num Layers \\
\midrule
Model~A (RIF) & 64  & 32 & 3 \\
Model~B & 64 & 32 & 2 \\
Model~C & 128 & 32 & 1 \\
Model~D & 64  & 32 & 3 \\
Model~E & 128 & 32 & 4 \\
Model~F & 64  & 32 & 1 \\
Model~A (RFB) & 128 & 32 & 1 \\
Model~A (INH) & 128 & 32 & 2 \\
Model~A (EMB) & 128 & 32 & 2 \\
\bottomrule
\end{tabular}
\end{center}
\end{table}   

\begin{table}[!htbp]
\begin{center} 
\caption{Comprehensive hyperparameter configurations for VAMP-Net. The table summarizes the optimized structural and learning parameters for Models A through F applied on RIF drug and model A apllied on RFB, INH and EMB drugs. This mapping includes the specific configurations for \textit{Path-2 (1D-CNN)}  and the Classification head.}
\label{table:hyperparams_pbch}
\begin{tabular}{@{}lccccc@{}}
\toprule
Model & Dropout & Activation & Conv Layers & Conv Kernel & LR \\
\midrule
Model~A (RIF) & 0.1128 & relu & 3 & 3 & 0.00137 \\
Model~B & 0.0962 & gelu & 2 & 5 & 0.00165 \\
Model~C & 0.3817 & gelu & 1 & 5 & 0.00130 \\
Model~D & 0.0205 & gelu & 3 & 3 & 0.00101 \\
Model~E & 0.0692 & relu & -- & -- & 0.00105 \\
Model~F & 0.3634 & gelu & -- & -- & 0.00154 \\
Model~A (RFB) & 0.2704 & relu & 4 & 5 & 0.00106 \\
Model~A (INH) & 0.326 & relu & 1 & 3 & 0.00141 \\
Model~A (EMB) & 0.326 & relu & 1 & 3 & 0.00141 \\
\bottomrule
\end{tabular}
\end{center} 
\end{table}   

\begin{figure}[H]
\centering
\begin{minipage}[b]{0.32\textwidth}
    \centering
    \includegraphics[width=\textwidth]{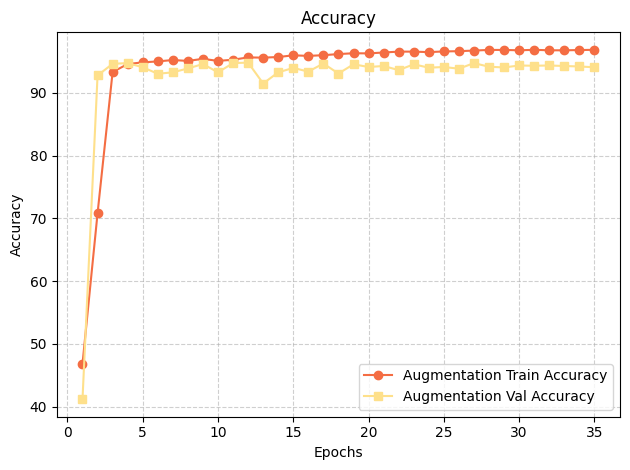}
\end{minipage}
\hfill
\begin{minipage}[b]{0.32\textwidth}
    \centering
    \includegraphics[width=\textwidth]{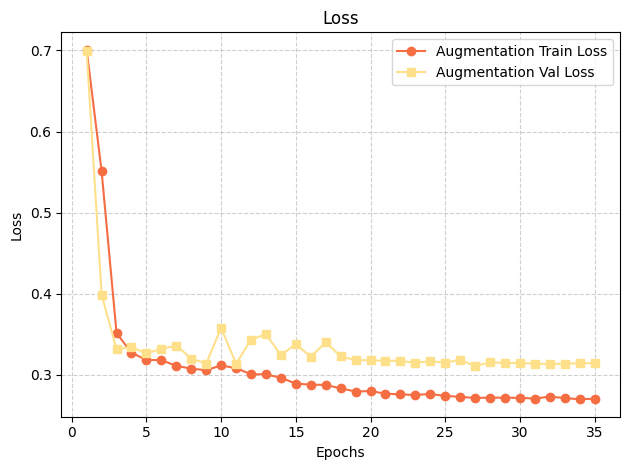}
\end{minipage}
\hfill
\begin{minipage}[b]{0.32\textwidth}
    \centering
    \includegraphics[width=\textwidth]{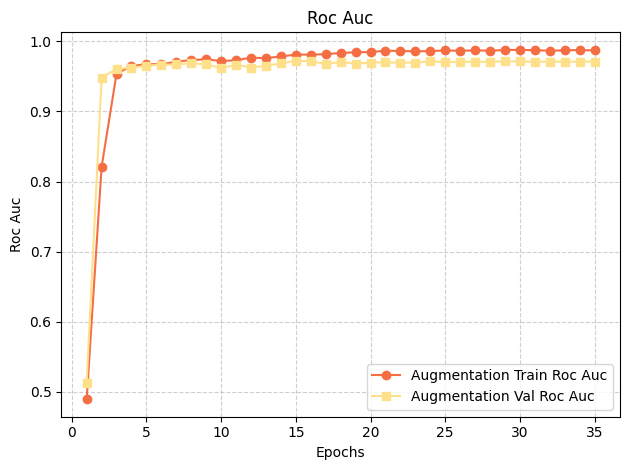}
\end{minipage}

\vspace{1em}

\begin{minipage}[b]{0.32\textwidth}
    \centering
    \includegraphics[width=\textwidth]{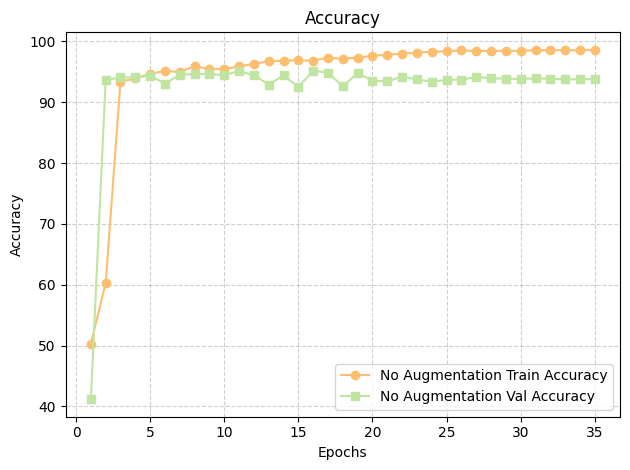}
\end{minipage}
\hfill
\begin{minipage}[b]{0.32\textwidth}
    \centering
    \includegraphics[width=\textwidth]{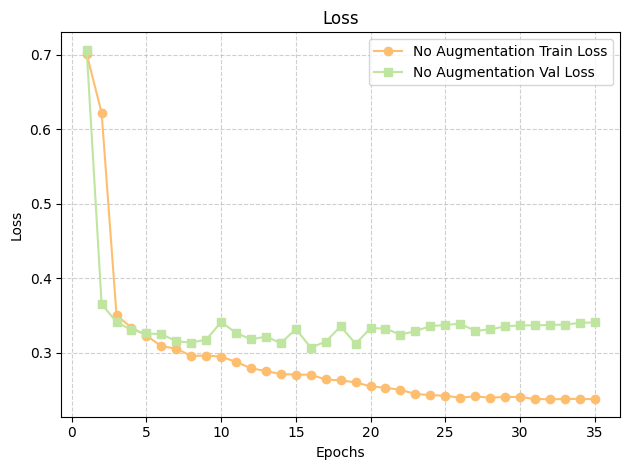}
\end{minipage}
\hfill
\begin{minipage}[b]{0.32\textwidth}
    \centering
    \includegraphics[width=\textwidth]{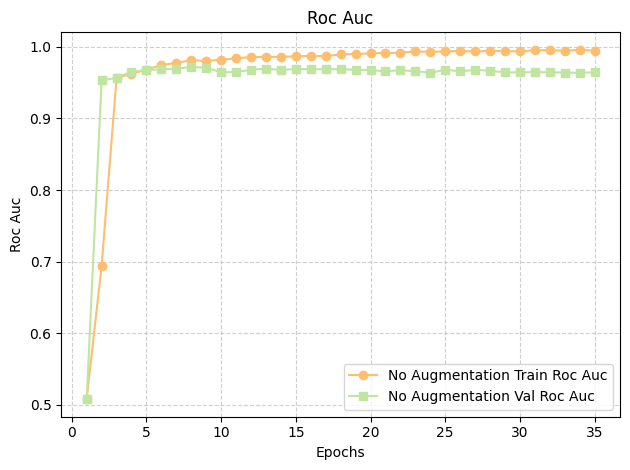}
\end{minipage}

\caption{Validation of permutation invariance via variant sequence shuffling. Comparative analysis of model performance under data augmentation (top row) and standard training (bottom row). The top panels demonstrate that variant shuffling does not degrade performance; rather, it acts as a regularizer that tightens the generalization gap. The stability of ROC-AUC and Accuracy across both rows, despite the randomized ordering of input mutations, validates the permutation-invariant inductive bias of the Set Attention Block (SAB), ensuring the model captures biological topology regardless of variant positioning.}
\label{fig:aug_vs_no_aug}
\end{figure}

The role of data augmentation, implemented via variant sequence shuffling, was scrutinized by comparing \textit{Model A} and \textit{Model B}. Shuffling leverages the Set Attention Block (SAB)'s inherent capacity for permutation invariance, ensuring that the model learns to identify critical genomic signatures independent of their relative position in the input sequence. As illustrated by the optimization trajectories in Figure~\ref{fig:aug_vs_no_aug}, \textit{Model B} exhibited superior training stability and a narrowed generalization gap, acting as an effective regularizer.
On the test set, \textit{Model-A} achieved the highest performance (ROC AUC: 0.969, Balanced Accuracy: 0.945). While \textit{Model-B}'s test performance was marginally lower (ROC AUC: 0.964, Balanced Accuracy: 0.941), this minor trade-off in peak performance resulted in a substantially more robust and stable model. This suggests that data augmentation effectively improves generalization and stability with only a negligible decrease in maximum predictive power.\\

The results summarized in Table \ref{tab:models_test_results_comparison} provide critical insights into the contribution of each modular component within the VAMP-Net (VN) network. A comparative analysis yields several key findings, the full VAMP-Net (VN) configuration, which incorporates attention masking and localized feature extraction, achieves the highest overall performance across most metrics, specifically reaching an Accuracy of 0.956 and an F1-score of 0.960. This confirms that the synergistic integration of the dual-path architecture and precise masking provides the most robust predictive signal for Rifampicin resistance. Notably, the removal of the attention mask ("No mask, no shuffle") does not result in a catastrophic loss of performance, maintaining a high ROC-AUC of 0.973. This suggests that the SAB mechanism is inherently capable of identifying informative tokens and effectively disregarding padding, even without explicit masking. However, the slightly lower Balanced Accuracy (0.942) compared to the baseline (0.944) indicates that masking still provides a marginal benefit in stabilizing predictions for minority classes. The exclusion of Path-2 ("No Path-2") consistently resulted in a performance decrement, with Balanced Accuracy dropping to its lowest point (0.930). This validates the hypothesis that Path-B captures complementary genomic information, likely broader contextual or inter-variant relationships—that a single-path transformer architecture might overlook.

\begin{table}[!htbp]
\caption{Quantitative performance comparison of VAMP-Net architectural variants on the RIF test set. The table evaluates the influence of attention masking, data augmentation (shuffling), and dual-path integration across four key metrics.}
\label{tab:models_test_results_comparison}
\centering
\small
\setlength{\tabcolsep}{6pt}
\begin{tabular}{lcccc}
\toprule
\textbf{Model variant} & \textbf{Accuracy} & \textbf{Bal. Acc.} & \textbf{F1-score} & \textbf{ROC-AUC} \\
\midrule
\textbf{A: VAMP-Net (VN)} & \textbf{0.956} & \textbf{0.944} & \textbf{0.960} & \textbf{0.970} \\

B: VN + shuffle & 0.945 & 0.941 & 0.950 & 0.963 \\
C: No mask, no shuffle & 0.944 & 0.942 & 0.953 & 0.973 \\
D: No mask + shuffle & 0.945 & 0.943 & 0.953 & 0.969 \\
E: No Path-2 + shuffle & 0.940 & 0.930 & 0.940 & 0.961 \\
F: No Path-2, no shuffle & 0.940 & 0.935 & 0.941 & 0.960 \\
\bottomrule
\end{tabular}
\end{table}


\subsubsection{VAMP-Net for anti-TB Drugs}

The predictive performance of VAMP-Net across four primary anti-TB drugs is detailed in Table~\ref{tab:seed-summary}. To rigorously evaluate model stability, we report metrics across two dimensions: validation stability (top panel) and final unseen test set efficacy (bottom panel). Across three independent experimental seeds, VAMP-Net demonstrated high architectural stability. Standard deviations for all primary metrics remained consistently low, with ROC-AUC variance specifically constrained to $< 0.007$. On the test set, INH achieved the highest overall performance, recording an accuracy of 0.945 $\pm$ 0.005 and a Matthews Correlation Coefficient (MCC) of 0.891 $\pm$ 0.010. RIF exhibited a similarly robust profile with an F1-score of 0.946 $\pm$ 0.007. For RFB, the model achieved its peak discriminative power with an ROC-AUC of 0.975 $\pm$ 0.001. A notable observation is the parity between Accuracy and Balanced Accuracy across most drugs (e.g., identical values of 0.945 for INH and a marginal 0.01 difference for EMB). Despite the inherent complexity of EMB resistance, VAMP-Net maintained the highest PR-AUC of the cohort at 0.977 $\pm$ 0.002. Collectively, the MCC values remained high across the spectrum (ranging from 0.614 to 0.891), indicating that the model maintains high-quality predictions across both resistant and susceptible classes.

\begin{table}[htbp]
\centering
\caption{Statistical performance profile of VAMP-Net across first-line and second-line anti-TB drugs. The top panel provides validation stability metrics (Mean $\pm$ SD) across three independent experimental seeds to assess model convergence. The bottom panel details the final predictive performance on the unseen test set. We report Accuracy, Balanced Accuracy, F1-score, Matthews Correlation Coefficient (MCC), and Area Under the ROC and Precision-Recall Curves (ROC-AUC, PR-AUC).}
\label{tab:seed-summary}
\footnotesize
\setlength{\tabcolsep}{4pt}
\begin{tabular}{lcccccc}
\toprule
Drug & Accuracy & Bal. Acc. & F1 & MCC & ROC-AUC & PR-AUC \\
\midrule
RIF & 0.937 $\pm$ 0.010 & 0.937 $\pm$ 0.013 & 0.946 $\pm$ 0.008 & 0.871 $\pm$ 0.022 & 0.967 $\pm$ 0.004 & 0.970 $\pm$ 0.005 \\
RFB & 0.925 $\pm$ 0.003 & 0.927 $\pm$ 0.002 & 0.924 $\pm$ 0.002 & 0.853 $\pm$ 0.003 & 0.971 $\pm$ 0.001 & 0.963 $\pm$ 0.002 \\
EMB & 0.839 $\pm$ 0.006 & 0.851 $\pm$ 0.007 & 0.892 $\pm$ 0.005 & 0.607 $\pm$ 0.007 & 0.915 $\pm$ 0.007 & 0.975 $\pm$ 0.002 \\
INH & 0.935 $\pm$ 0.007 & 0.935 $\pm$ 0.007 & 0.936 $\pm$ 0.007 & 0.870 $\pm$ 0.014 & 0.964 $\pm$ 0.000 & 0.953 $\pm$ 0.002 \\
\bottomrule
\end{tabular}
\vspace{6pt}
\textbf{Test metrics:}\\[2pt]
\begin{tabular}{lcccccc}
\toprule
Drug & Accuracy & Bal. Acc. & F1 & MCC & ROC-AUC & PR-AUC \\
\midrule
RIF & 0.937 $\pm$ 0.009 & 0.935 $\pm$ 0.013 & 0.946 $\pm$ 0.007 & 0.870 $\pm$ 0.019 & 0.968 $\pm$ 0.006 & 0.970 $\pm$ 0.009 \\
RFB & 0.935 $\pm$ 0.008 & 0.937 $\pm$ 0.007 & 0.934 $\pm$ 0.006 & 0.874 $\pm$ 0.013 & \textbf{0.975 $\pm$ 0.001} & 0.960 $\pm$ 0.003 \\
EMB & 0.843 $\pm$ 0.014 & 0.853 $\pm$ 0.009 & 0.895 $\pm$ 0.010 & 0.614 $\pm$ 0.022 & 0.920 $\pm$ 0.004 & \textbf{0.977 $\pm$ 0.002} \\
INH & \textbf{0.945 $\pm$ 0.005} & \textbf{0.945 $\pm$ 0.005} & \textbf{0.945 $\pm$ 0.005} & \textbf{0.891 $\pm$ 0.010} & 0.970 $\pm$ 0.002 & 0.961 $\pm$ 0.004 \\
\bottomrule
\end{tabular}
\end{table}

\begin{figure}[H]
\centering
\includegraphics[width=1\linewidth]{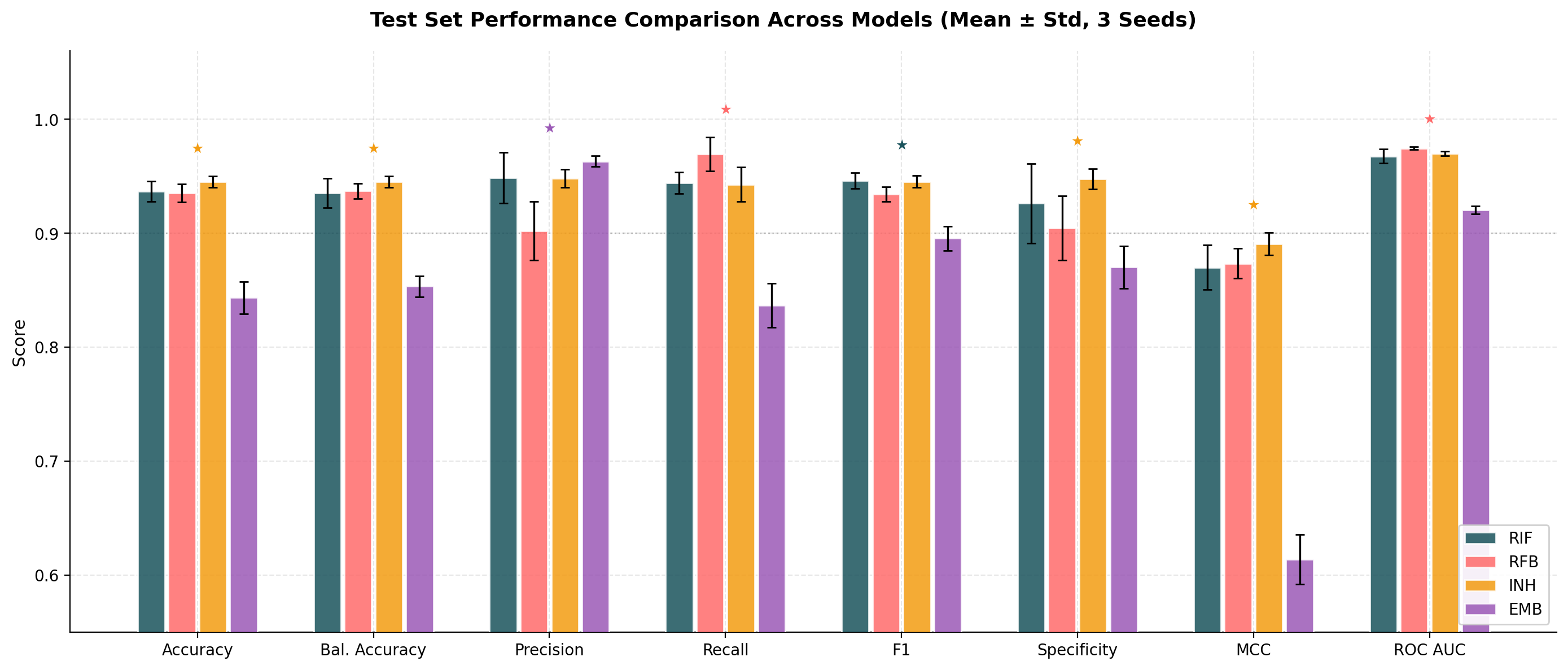}
\caption{Comparative test set performance across first-line and second-line anti-TB drugs. The bar chart illustrates the mean performance metrics across three independent experimental seeds for RIF, RFB, INH, and EMB. Error bars represent the standard deviation ($\pm$ SD), indicating high model stability across different initializations. Star markers ($\star$) denote the peak performance achieved for specific drug-metric pairs.}
\label{fig:a1}
\end{figure}

To evaluate the clinical readiness of VAMP-Net, we conducted a multi-dimensional analysis focusing on predictive accuracy, computational throughput, and stochastic stability across four major anti-TB drugs. As illustrated in Figure~\ref{fig:a1}, VAMP-Net demonstrated high discriminative performance across the cohort. The model achieved peak ROC-AUC values exceeding $0.96$ for RIF, RFB, and INH, with INH specifically reaching a mean Accuracy and Balanced Accuracy of $0.945 \pm 0.005$. While EMB presented the most challenging classification task, indicated by a lower Matthews Correlation Coefficient (MCC) of $0.614$, it maintained a superior Precision-Recall AUC ($0.977$), suggesting that the model remains highly effective at identifying true resistant isolates despite significant class imbalance.\\

\begin{figure}[H]
\centering
\includegraphics[width=1\linewidth]{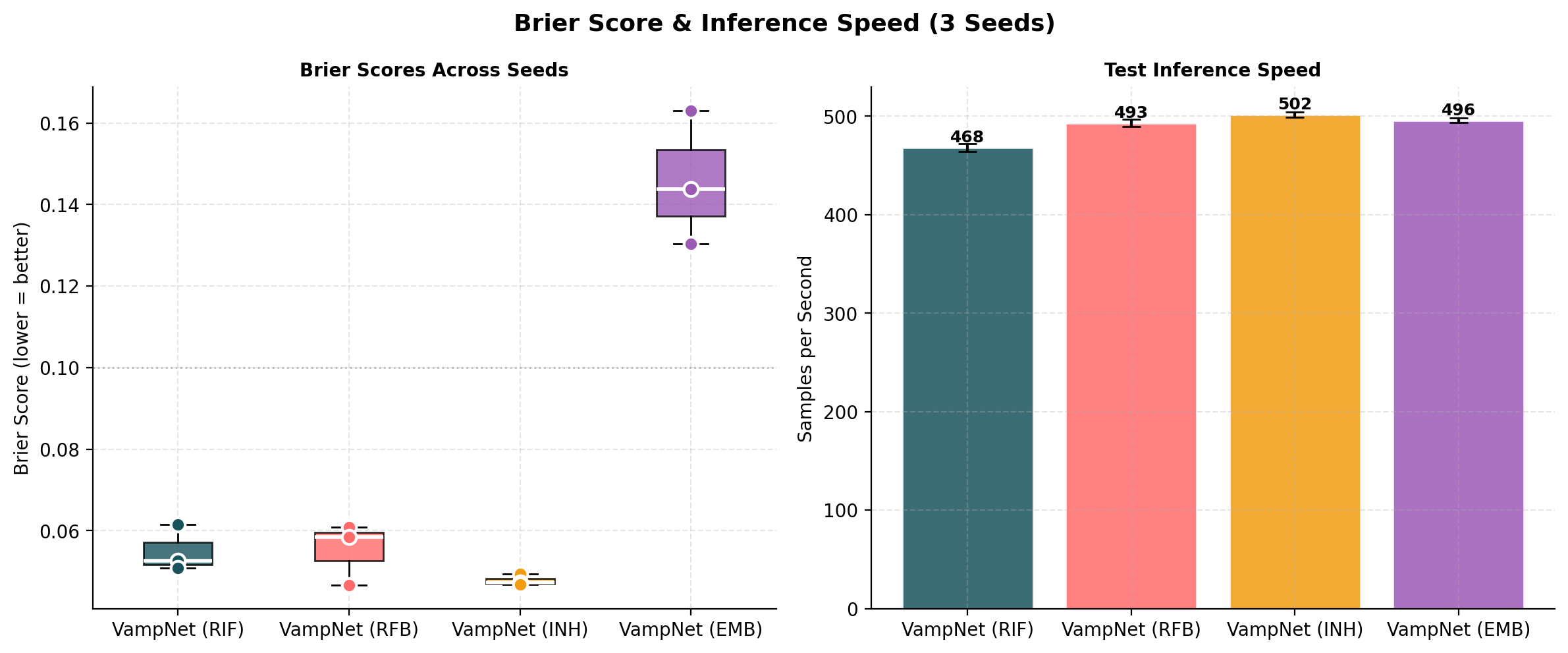}
\caption{Computational efficiency and calibration across targeted anti-TB drugs. Left: Boxplots illustrate the distribution of Brier scores across three independent seeds, where a lower score indicates superior model calibration and probabilistic accuracy.  Right: Throughput analysis measured in samples per second during inference. VAMP-Net maintains a consistent high-speed processing rate across all drugs, averaging $\sim$490 samples/s. Error bars represent the standard deviation ($\pm$ SD) across three experimental runs.}
\label{fig:a2}
\end{figure}

\begin{figure}[H]
\centering
\includegraphics[width=1\linewidth]{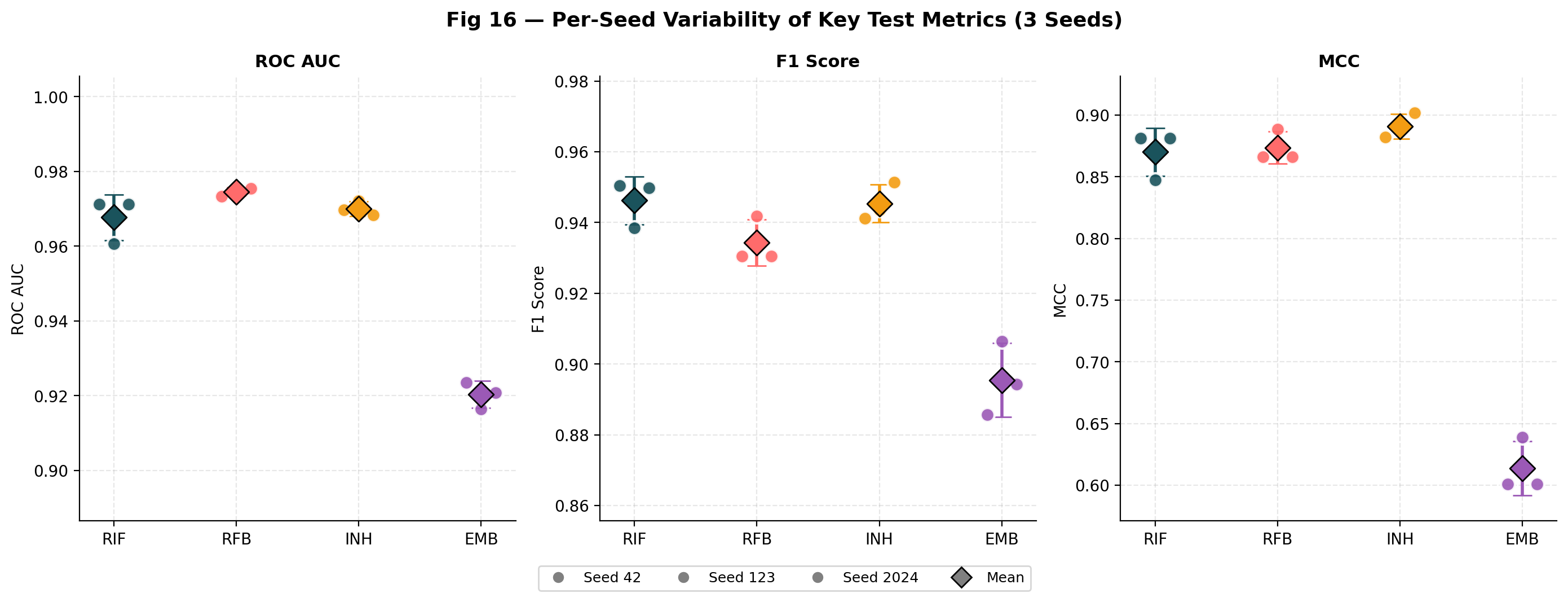}
\caption{Per-seed variability of key performance metrics across three independent experimental runs. The scatter-boxplots illustrate the distribution of ROC-AUC (left), F1-score (center), and Matthews Correlation Coefficient (MCC, right) for RIF, RFB, INH, and EMB. Individual points represent specific stochastic seeds (Seed 42, Seed 123, and Seed 2024), while the diamond marker ($\diamond$) denotes the arithmetic mean. Error bars indicate the range of variability across runs. The high degree of clustering around the mean for RIF, RFB, and INH demonstrates the architectural stability of VAMP-Net. While EMB exhibits slightly higher dispersion, reflecting its greater biological and dataset complexity, the consistent performance across different initializations underscores the robustness of the Set Attention mechanism in capturing significant resistance signatures.}
\label{fig:a3}
\end{figure}

The calibration of our model’s probabilistic outputs was assessed using the Brier Score (Figure \ref{fig:a2} Left). VAMP-Net exhibited high calibration stability for INH ($\sim$0.05) and moderate calibration for RIF and RFB. The higher Brier score for EMB ($\sim$0.14) aligns with its lower MCC, further highlighting the biological complexity inherent in Ethambutol resistance pathways. Simultaneously, inference speed was evaluated to determine suitability for clinical deployment. VAMP-Net maintained a high-throughput processing rate of $\sim$490 samples per second across all drugs ( Figure \ref{fig:a2} Right), ensuring near-instantaneous diagnostics even for large genomic surveillance datasets. To ensure the reported metrics were not artifacts of favorable initialization, we measured per-seed variability (Figure \ref{fig:a3}). The results show tight clustering around the mean for ROC-AUC, F1-score, and MCC across three independent seeds. The negligible variance in RIF and INH performance underscores the architectural robustness of our attention-based framework.

\subsection{Comparative Analysis}

\subsubsection{Comparative Performance Against Baseline Architectures}

We benchmarked our approach against the traditional binary encoding paradigm prevalent in antimicrobial resistance (AMR) and clinical genomics literature \citep{sakagianni2024data}. In this conventional framework, genomic data is represented as a high-dimensional sparse matrix $\mathbf{X} \in \{0,1\}^{N \times V}$, where $X_{i,j}$ denotes the presence or absence of variant $j$ in sample $i$. While effective in certain contexts, this representation suffers from the curse of dimensionality, necessitating significant computational resources or aggressive feature selection, such as PCA or statistical pruning, to manage the sparse feature space \citep{reshetnikov2025feature, pikalyova2024predicting, xu2025machine}. To ensure a fair comparison, we performed feature selection using a Chi-squared test ($\alpha = 0.001$) to retain significantly associated variants, resulting in a reduced set of 11,567 features ($\sim$36.1\% reduction) used to train VAMP-Net and competitive Multi-Layer Perceptron (MLP) and Convolutional Neural Network (CNN) baselines applied on \textit{RIF} drug. The optimazed hyperparameter configurations for MLP and CNN are presented in Tables \ref{table:hyperparams_mlp}, and \ref{table:hyperparams_cnn} respectively.\\

\begin{table}[!htbp]
\begin{center} 
\caption{ Optimized hyperparameter configuration for the MLP baseline. The table outlines the best-performing structural parameters  determined via empirical tuning on Rifampicin (RIF) resistance prediction.This specific configuration is referred to as Model G}
\label{table:hyperparams_mlp}
\begin{tabular}{@{}lccc@{}}
\toprule
Model & Hidden Dims & Dropout & Activation \\
\midrule
Model~G & [1024, 128, 128] & 0.1315 & gelu \\
\bottomrule
\end{tabular}
\end{center} 
\end{table}

\begin{table}[!htbp]
\begin{center} 
\caption{Optimized hyperparameter configuration for the CNN baseline. The table outlines the best-performing structural parameters  determined via empirical tuning on Rifampicin (RIF) resistance prediction.This specific configuration is referred to as Model H}
\label{table:hyperparams_cnn}
\begin{tabular}{@{}lccccccc@{}}
\toprule
Model & Dropout & Activation & Conv Layers & Filters & Kernel Sizes & Pool Sizes \\
\midrule
Model~H & 0.1487 & relu & 3 & [96, 128, 256] & [5, 9, 2] & [3, 3, 3] \\
\bottomrule
\end{tabular}
\end{center} 
\end{table}

\begin{figure}[H]
\centering
\includegraphics[width=1\linewidth]{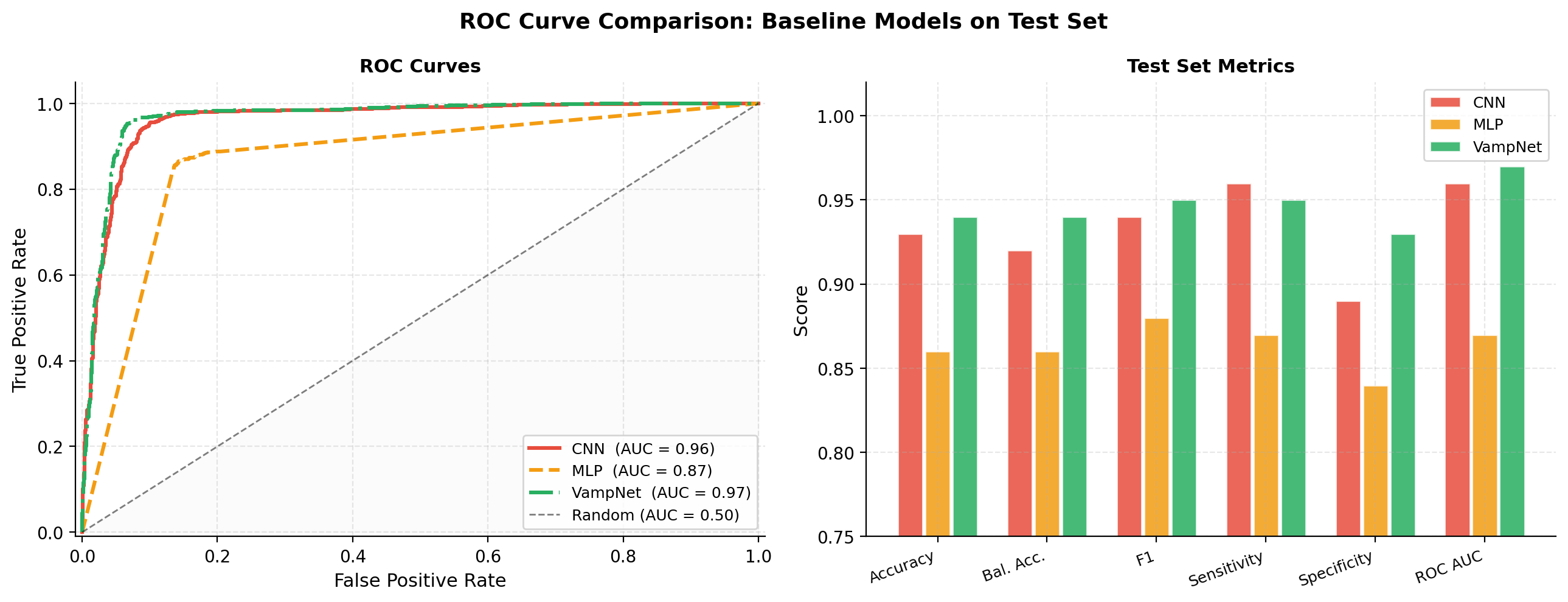}
\caption{Comparative performance of VAMP-Net against baseline architectures on the test set. (Left) Receiver Operating Characteristic (ROC) curves demonstrating the trade-off between true positive and false positive rates. (Right) Detailed test set metrics across six performance indicators.}
\label{fig:roc_auc_traditional_vs_sab}
\end{figure}

\begin{figure}[H]
\centering
\includegraphics[width=1\linewidth]{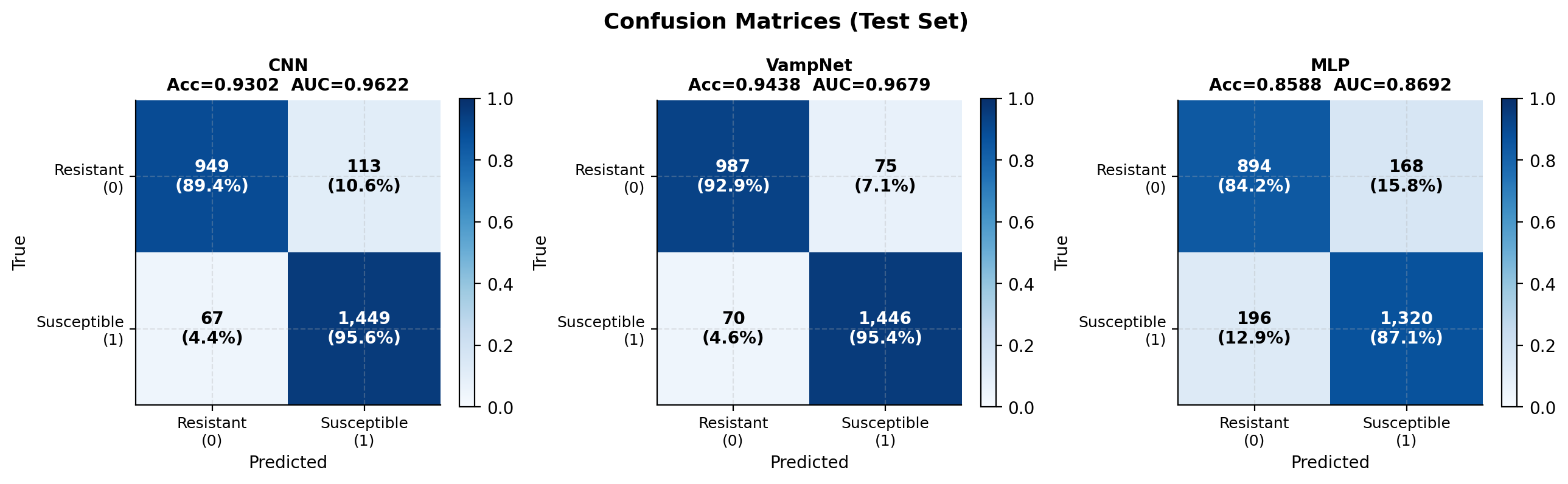}
\caption{Comparative confusion matrix analysis on the unseen test set. This figure provides a head-to-head performance breakdown between the proposed VAMP-Net architecture and two baseline models: a Convolutional Neural Network (CNN) and a Multi-Layer Perceptron (MLP).}
\label{fig:cn_traditional_vs_sab}
\end{figure}

As illustrated in Figure \ref{fig:roc_auc_traditional_vs_sab}, VAMP-Net consistently establishes a new performance ceiling across all evaluated metrics. The ROC curve analysis  reveals that while the CNN achieves a competitive AUC of 0.96, VAMP-Net reaches a superior 0.97, demonstrating a more robust trade-off between sensitivity and specificity. The MLP, lacking specialized spatial or attention-based feature extraction, lags significantly with an AUC of 0.87. The granularity of this superiority is further evidenced by the confusion matrices on the unseen test set (Figure \ref{fig:cn_traditional_vs_sab}). VAMP-Net correctly identifies 92.9\% of resistant isolates, a notable improvement over the CNN (89.4\%) and the MLP (84.2\%).  Crucially, as shown in Table \ref{tab:vampnet_condensed}, the performance gains are statistically validated; VAMP-Net achieved a significant improvement in Accuracy over the CNN 
($+1.35\%$, $p < 0.001$) and the MLP ($+8.48\%$, $p < 0.001$). The 95\% Confidence Intervals for VAMP-Net remain narrow (e.g., Accuracy: $[0.9383, 0.9554]$), underscoring the model's high reliability and precision in a clinical diagnostic context.

\begin{table}[htbp]
\centering
\small 
\setlength{\tabcolsep}{4pt} 
\caption{VAMP-Net overall performance and statistical significance comparison against baselines.}
\label{tab:vampnet_condensed}
\begin{tabular*}{\columnwidth}{@{\extracolsep{\fill}}lccc}
\toprule
\textbf{Metric} & \textbf{Value} & \textbf{95\% CI} & \textbf{p-value (vs Baselines)} \\
\midrule
Accuracy & 0.9474 & [0.9383, 0.9554] & $< 0.001^{**}$ \\
F1-score & 0.9555 & [0.9476, 0.9627] & $< 0.001^{**}$ \\
ROC-AUC  & 0.9691 & [0.9612, 0.9762] & 0.074 \\
\bottomrule
\addlinespace
\multicolumn{4}{l}{\scriptsize $^{**}$Significant at $p < 0.001$ relative to CNN and MLP baselines.}
\end{tabular*}
\end{table}

\subsubsection{Comparative Evaluation of Fusion Paradigms}
\label{subsec:fusion_comparison}

The architectural efficacy of VAMP-Net’s late-stage adaptive fusion was benchmarked against a traditional Early Fusion baseline, applied to \textit{RIF} drug,  to quantify the impact of integration timing on model performance. As summarized in Table \ref{tab:early-late-fusion}, VAMP-Net significantly outperformed the Early Fusion paradigm across all evaluated metrics. Most notably, VAMP-Net achieved transformative gains in Specificity (+48.8\%), Balanced Accuracy (+32.6\%), and ROC-AUC (+30.9\%) compared to the early-integration approach. The Early Fusion model exhibited a profound performance asymmetry, demonstrating high Sensitivity (0.805) but catastrophic Specificity (0.425). This imbalance suggests a systematic failure to distinguish true genomic resistance markers from stochastic noise. In contrast, VAMP-Net maintained a harmonized performance profile, achieving a Recall of 0.968 and a Specificity of 0.913.\\

The empirical results indicate that forcing inter-modality interaction at a premature stage—prior to dedicated contextual modeling—is suboptimal for complex genomic tasks. In the Early Fusion setting, token-level embeddings and structured features are concatenated and projected before entering the Set Attention Blocks (SABs). Our findings suggest that this approach leads to the propagation of unrefined signal noise, which the subsequent attention mechanisms struggle to resolve, as evidenced by the severe specificity deficit. VAMP-Net’s superiority is rooted in its late-stage adaptive fusion philosophy. By postponing integration until after modality-specific encoding, each pathway is permitted to develop robust internal representations. This allows the SAB encoder to model inter-variant dependencies within a more stable feature space.

\begin{table}[htbp]
\centering
\caption{Comparative analysis of fusion architectural paradigms. This table benchmarks the proposed VAMP-Net multi-path integration (late-stage adaptive fusion) against a traditional Early Fusion baseline.}
\label{tab:early-late-fusion}
\small
\setlength{\tabcolsep}{8pt}
\renewcommand{\arraystretch}{1.15}
\begin{tabular}{lcc}
\toprule
\textbf{Metric} & \textbf{Early Fusion} & \textbf{VAMP-Net Fusion} \\
\midrule
Accuracy & 0.649 & \textbf{0.946} \\
F1-score & 0.729 & \textbf{0.955} \\
Sensitivity (Recall) & 0.805 & \textbf{0.968} \\
Specificity & 0.425 & \textbf{0.913} \\
Precision (PPV) & 0.667 & \textbf{0.941} \\
NPV & 0.605 & \textbf{0.953} \\
Balanced Accuracy & 0.615 & \textbf{0.941} \\
ROC-AUC & 0.655 & \textbf{0.964} \\
\bottomrule
\end{tabular}
\end{table}


\subsubsection{Comparative SOTA}

To provide a more faithful comparison, we separate prior work by paper and retain only the metrics explicitly reported in each study. For every drug shared with our benchmark, we compare the published result against the best-performing VAMP-Net seed for that drug on our validation protocol. Because these studies use different cohorts, split strategies, and prediction formulations, the following tables should be interpreted as paper-specific contextual comparisons rather than direct leaderboard claims.

\begin{table*}[t]
\centering
\small
\caption{Paper-specific comparison with the XGBoost study in \textit{Diagnostics} \citep{paredes2025predicting}. The original paper reports Sensitivity, Specificity, Precision, F1-score, and Accuracy for EMB, INH, and RIF.}
\label{tab:sota_diagnostics}
\resizebox{\textwidth}{!}{
\begin{tabular}{lccccccccccc}
\hline
\textbf{Drug} & \textbf{Our Seed} & \textbf{Sens. (Paper)} & \textbf{Sens. (Ours)} & \textbf{Spec. (Paper)} & \textbf{Spec. (Ours)} & \textbf{Prec. (Paper)} & \textbf{Prec. (Ours)} & \textbf{F1 (Paper)} & \textbf{F1 (Ours)} & \textbf{Acc. (Paper)} & \textbf{Acc. (Ours)} \\
\hline
EMB & 42 & 0.97 & 0.8166 & 0.97 & 0.8885 & 0.89 & 0.9673 & 0.93 & 0.8856 & 0.97 & 0.8309 \\
INH & 42 & 0.90 & 0.9382 & 0.99 & 0.9439 & 0.97 & 0.9441 & 0.94 & 0.9411 & 0.97 & 0.9410 \\
RIF & 2024 & 0.94 & 0.9360 & 0.96 & 0.9501 & 0.90 & 0.9640 & 0.92 & 0.9498 & 0.96 & 0.9418 \\
\hline
\end{tabular}
}
\end{table*}

Compared with the XGBoost results reported in \textit{Diagnostics}, VAMP-Net shows a more favorable precision profile across all shared drugs, with particularly strong gains for EMB and RIF, indicating that the proposed architecture is more conservative in assigning resistant calls while preserving competitive sensitivity. The main trade-off appears for EMB, where the prior study reports higher sensitivity and accuracy, suggesting that EMB remains the most challenging phenotype in our setting and may be more sensitive to cohort composition and decision-threshold effects.

\begin{table*}[t]
\centering
\small
\caption{Paper-specific comparison with the MIC-prediction machine learning system reported in \textit{PLOS Computational Biology} \citep{cryptic2024quantitative}. Only metrics shared with our evaluation are shown; the paper reports PPV, which is reported here as the published precision-equivalent quantity.}
\label{tab:sota_plos}
\resizebox{\textwidth}{!}{
\begin{tabular}{lccccccccccc}
\hline
\textbf{Drug} & \textbf{Our Seed} & \textbf{AUC (Paper)} & \textbf{AUC (Ours)} & \textbf{Sens. (Paper)} & \textbf{Sens. (Ours)} & \textbf{Spec. (Paper)} & \textbf{Spec. (Ours)} & \textbf{PPV/Prec. (Paper)} & \textbf{Prec. (Ours)} & \textbf{Acc. (Paper)} & \textbf{Acc. (Ours)} \\
\hline
EMB & 42 & 0.965 & 0.9236 & 0.931 & 0.8166 & 0.894 & 0.8885 & 0.704 & 0.9673 & 0.902 & 0.8309 \\
INH & 42 & 0.983 & 0.9696 & 0.950 & 0.9382 & 0.981 & 0.9439 & 0.980 & 0.9441 & 0.966 & 0.9410 \\
RIF & 2024 & 0.992 & 0.9712 & 0.971 & 0.9360 & 0.973 & 0.9501 & 0.962 & 0.9640 & 0.972 & 0.9418 \\
RFB & 2024 & 0.988 & 0.9733 & 0.955 & 0.9782 & 0.957 & 0.8882 & 0.918 & 0.8874 & 0.956 & 0.9309 \\
\hline
\end{tabular}
}
\end{table*}

Relative to the MIC-oriented model from \textit{PLOS Computational Biology}, VAMP-Net remains competitive but reveals a mixed pattern. Our model achieves its closest agreement on INH and RFB, while the largest performance gaps are observed for EMB and RIF, especially in AUC and accuracy. This suggests that VAMP-Net is strongest when technical-confidence recalibration contributes substantially to the final decision, but that MIC-prediction frameworks trained under different targets and evaluation designs can retain an advantage on some first-line drugs.

\begin{table*}[t]
\centering
\small
\caption{Paper-specific comparison with LLMTB, the BERT-based large language model reported in \textit{Bioinformatics} \citep{testagrose2025leveraging}. The paper reports positive-class F1-scores on the shared drugs.}
\label{tab:sota_llmtb}
\resizebox{0.72\textwidth}{!}{
\begin{tabular}{lccc}
\hline
\textbf{Drug} & \textbf{Our Seed} & \textbf{F1 (LLMTB)} & \textbf{F1 (Ours)} \\
\hline
EMB & 42 & 0.8273 & 0.8856 \\
INH & 42 & 0.9361 & 0.9411 \\
RIF & 2024 & 0.9094 & 0.9498 \\
RFB & 2024 & 0.8913 & 0.9306 \\
\hline
\end{tabular}
}
\end{table*}

Against LLMTB, VAMP-Net improves the reported F1-score for all four shared drugs, with the clearest gains on EMB, RIF, and RFB. This pattern supports the value of combining permutation-invariant biological modeling with an explicit quality-aware branch, rather than relying solely on sequence-style language representations. The smaller margin on INH further suggests that both architectures are already highly effective for this phenotype, leaving less headroom for separation.



\subsection{Path-2 Contribution and Case Studies: }

To better understand how the quality-aware Path-2 branch influences final predictions, we inspected representative test cases in which the fused model either \textit{corrected} an overconfident Path-1 decision or \textit{amplified} a borderline but correct Path-1 prediction. Table~\ref{tab:path2_case_studies} summarizes these case studies using the resistance probability predicted by the variant-only pathway (Path-1) and by the full fused model. We additionally report the probability shift $\Delta P = P_{\mathrm{Full}}(\mathrm{Resistant}) - P_{\mathrm{Path\mbox{-}1}}(\mathrm{Resistant})$, together with summary statistics of the learned fusion gate. These examples show that the VCF-aware Path-2 branch is not merely an auxiliary source of features, but an active recalibration mechanism for Path-1’s variant-based decision. In the corrected cases, Path-1 predicted resistance with very high confidence ($P(R)\approx 0.91$), yet the fused model reduced this probability to $0.04$--$0.06$, corresponding to a large decision shift of roughly $-0.86$. This is not a marginal adjustment; it is a strong reversal of an initially overconfident prediction.\\

\begin{table*}[htbp]
\centering
\caption{\textbf{Ablation case studies of VCF-aware Path-2 contribution.} All samples feature a variant count of 743. ``Corrected'' denotes cases where Path-B reversed an incorrect Path-1 prediction, while ``Amplified'' denotes Path-2 reinforcing a correct class by shifting borderline predictions away from the decision boundary.}
\label{tab:path2_case_studies}
\small
\setlength{\tabcolsep}{4pt} 
\renewcommand{\arraystretch}{1.3}
\begin{tabularx}{\textwidth}{lcccc X} 
\toprule
\textbf{True Label} & \textbf{Path-1 $P(R)$} & \textbf{Full $P(R)$} & \textbf{$\Delta P$} & \textbf{Effect} & \textbf{Principal Explanation} \\
\midrule
Susceptible & 0.9190 & 0.0521 & $-0.867$ & Corrected & High FRS but low genotype-confidence and depth support made Path-1's resistance call unreliable. \\

Susceptible & 0.9049 & 0.0438 & $-0.861$ & Corrected & Path-A overcalled resistance; Path-2 utilized VCF-support signals to downweight the prediction. \\

Susceptible & 0.9185 & 0.0592 & $-0.859$ & Corrected & Low-confidence and low-depth evidence triggered the fused model to reverse the initial resistant call. \\

Susceptible & 0.4579 & 0.0774 & $-0.381$ & Amplified & Borderline sample pushed further into the susceptible class, increasing overall model confidence. \\

Resistant   & 0.5491 & 0.9028 & $+0.354$ & Amplified & Combined variant identity with consistent support features to strengthen a true resistant call. \\

Resistant   & 0.5495 & 0.8826 & $+0.333$ & Amplified & Reinforced the resistant signal, moving a borderline Path-1 output safely away from the decision boundary. \\
\bottomrule
\end{tabularx}
\end{table*}

\begin{figure}[htbp]
\centering
\includegraphics[width=0.95\linewidth,height=0.72\textheight,keepaspectratio]{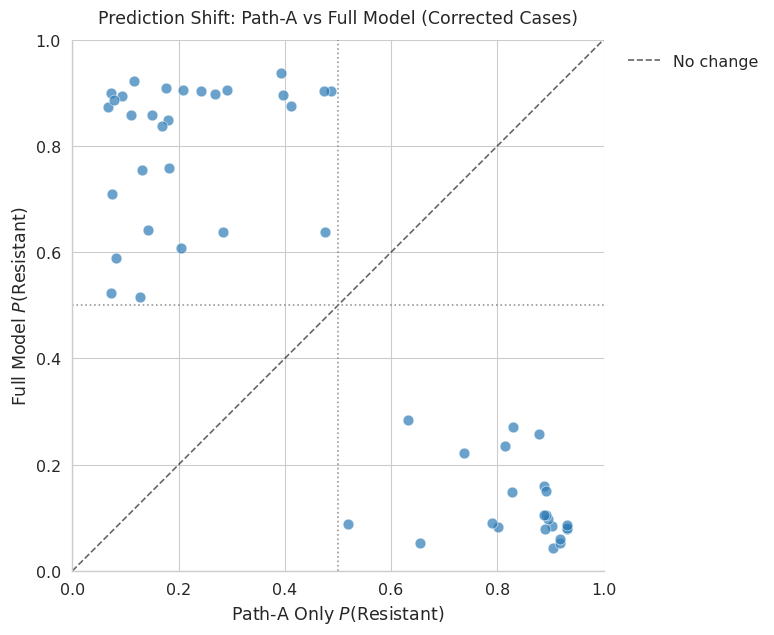}
\caption{Probabilistic recalibration of resistance calls through multi-path fusion. The scatter plot illustrates the shift in resistance probability $P(\text{Resistant})$ when transitioning from the variant-only model (Path-1) to the fully integrated VAMP-Net architecture (Full Model). Points in the lower-right quadrant represent "Corrected" cases, where Path-1 initially assigned high resistance probabilities ($>0.5$) that were subsequently reversed by Path-2's quality-aware branch to confident susceptible calls ($<0.2$). Conversely, points in the upper-left quadrant denote "Amplified" cases, where Path-2 reinforced borderline or incorrect susceptible predictions into high-confidence resistance calls. The significant deviation from the "No change" diagonal underscores Path2's role as a critical reliability filter that identifies and corrects overconfident predictions based on underlying sequencing evidence.}
\label{fig:path-2_cont}
\end{figure}

Inspection of the associated feature profiles clarifies why this occurs. In these corrected cases, \textit{FRS} remained high, indicating that the called alleles were internally supported by reads, but the broader VCF context, particularly \textit{GT\_CONF\_PERCENTILE}, depth-related features, and filtered depth support, remained weak. Thus, although Path-1 detected a mutation pattern that appeared resistance-like, Path-2 identified insufficient technical evidence to trust that pattern as a robust resistance signal. In this sense, Path-2 behaves as a reliability-aware filter: variant presence alone is not treated as sufficient, and the model instead asks whether the surrounding sequencing evidence is strong and consistent enough to justify a resistant call. The amplified cases demonstrate the complementary behavior. When Path-1 produced borderline predictions, Path-2 was able to push the fused output farther toward the correct class. For truly resistant isolates, the full model increased the resistance probability from about $0.55$ to $0.88$--$0.90$, showing that the quality-aware branch does not only suppress unreliable signals, but can also strengthen biologically meaningful signals when the VCF support is coherent. Overall, these case studies show that the fused model can correct Path-1 false positives by incorporating genotype confidence, depth, and coverage-derived evidence, thereby reducing spurious resistance probabilities by as much as $0.87$ and yielding more reliable susceptibility calls. Figure \ref{fig:path-2_cont} provides a visual quantification of the "corrective" and "amplificatory" influence of the Path-2 quality-aware branch. By plotting the resistance probability of Path-1 against the final model output, we can observe the specific instances where sequencing metadata fundamentally altered the diagnostic outcome.

\subsection{Multi-Path Network Interpretation and Epistasis}

This section prensents the results of the Dual-Path Interpretability Network and leverages the insights to dissect the model's decision-making process for \textit{RIF}, \textit{RFB}, \textit{INH} and \textit{EMB} resistance prediction. Our analysis demonstrates that the two architectural pathways within \textit{VAMP-Net} provide distinct yet complementary perspectives on the underlying mechanisms of drug resistance. Specifically, Path-1 (Set Attention Transformer) successfully identified and quantified variant-level importance and epistatic dependencies, thus elucidating the core genetic drivers. In contrast, Path-2 (Quality-Aware CNN) focused on quantifying the influence of technical confidence metrics, revealing how the model performs noise regularization and feature calibration. By integrating the findings from both streams, our network provides a comprehensive and auditable explanation of the model’s decision process, both biological and technical,  that drives the final drug resistance classification.

\subsubsection {Mechanistic Interpretation and Variant Attribution}

To quantify the individual genetic contributions to drug resistance, we applied the Integrated Gradients (IG) method to the $\text{SAB}$ pathway. As detailed in Table \ref{tab:combined_variants}, the resulting importance scores establish a hierarchical signature that validates the network’s ability to autonomously rediscover established molecular pathways while proposing high-impact novel candidates.

\begin{table}[htbp]
\centering
\caption{Top-ranked genomic variants and associated genes identified by VAMP-Net. The table lists high-importance variants prioritized by the model’s attention mechanism for RIF, RFB, INH and EMB. While the model successfully recovers established resistance determinants in the rpoB, katG, and ethA genes, it also identifies several "novel" high-impact loci in genes such as ponA1, pknB, and mmpL5.}
\label{tab:combined_variants}
\small
\begin{tabular}{@{}llcl@{}}
\toprule
\textbf{Drug} & \textbf{Variant} & \textbf{Importance Score} & \textbf{Gene} \\ \midrule
\textit{RIF} & 2155162\_ATGCCGC$>$ACGCCGT & 0.446 & \textit{katG} \\
 & 2338202\_CAC$>$CCAC & 0.308 & \textit{ponA1} (novel) \\
 & 424791\_T$>$TAT & 0.273 & \textit{pknB} (novel) \\
 & 1849861\_G$>$T & 0.207 & \textit{ethA} \\
 & 761139\_C$>$A & 0.178 & \textit{rpoB} \\
 & 761101\_A$>$T & 0.116 & \textit{rpoB} \\ \midrule
\textit{RFB} & 761140\_A$>$T & 6.896 & \textit{rpoB} \\
 & 2944085\_C$>$T & 4.947 & \textit{mmpL5} (novel) \\
 & 893702\_GAC...CCG & 4.213 & \textit{lpqS} (novel) \\
 & 761139\_C$>$A & 3.955 & \textit{rpoB} \\
 & 3640407\_G$>$A & 3.532 & \textit{ppsD} (novel) \\
 & 3545402\_A$>$G & 3.527 & \textit{fadD26} region (novel) \\
 & 1262023\_C$>$T & 2.885 & \textit{pks13} (novel) \\
\midrule
\textit{EMB} & 740348\_TCGCGCGAC$>$T & 0.756 & \textit{lipG} (novel) \\
 & 764509\_G$>$C & 0.638 & \textit{rpoC} (novel) \\
 & 740383\_G$>$T & 0.559 & \textit{lipG} (novel) \\
 & 2435103\_T$>$C & 0.536 & \textit{idsA2} (novel) \\
 & 4269298\_A$>$G & 0.180 & \textit{ubiA} (known) \\
 & 4247429\_A$>$C & 0.062 & \textit{embB} (known) \\
 & 4247429\_A$>$T & 0.055 & \textit{embB} (known) \\
 & 4248003\_A$>$G & 0.046 & \textit{embB} (known) \\
 & 2603465\_C$>$G & 0.502 & intergenic (regulatory candidate) \\
 & 2113033\_G$>$C & 0.363 & intergenic (regulatory candidate) \\
 \midrule
\textit{INH} & 2760136\_TCC...ATA$>$TTC...ATG & 8.195 & \textit{mmuM} (novel) \\
 & 4366246\_GCC...CG$>$GGC...GCC & 0.763 & \textit{eccA2} (novel) \\
 
 & 575699\_G$>$A & 0.085 & \textit{mshA} (known) \\
 & 413294\_ACCC...$>$A & 0.097 & \textit{iniC} (known) \\

 & 1503392\_ACATG...GGTTG$>$A & 0.527 & intergenic (regulatory candidate) \\
 & 1113698\_CACG...GATGA$>$C & 0.502 & intergenic (regulatory candidate) \\
\bottomrule
\end{tabular}
\end{table}

VAMP-Net assigned primary predictive weight to canonical mutations within the RRDR of the $rpoB$ gene, precisely aligning with WHO-validated markers for Rifampicin (RIF) and Rifabutin (RFB), where the top-ranked RFB variant 761140\_A$>$T achieved a dominant score of 6.896. Notably, the model extracted complex Multi-Drug Resistance (MDR) signatures from single-drug tasks; for RIF, the highest importance score (0.446) was attributed to the Isoniazid-associated $katG$ variant, while for Ethambutol (EMB), the model correctly prioritized the $rpoC$ variant (0.638). This confirms that VAMP-Net leverages co-selection patterns and compensatory mechanisms inherent in MDR-TB lineages to enhance its predictive confidence.A critical outcome of the attribution analysis is the identification of high-scoring variants in uncharacterized regions, most significantly the \textit{mmuM} variant for Isoniazid (INH), which attained the study's highest overall importance score of 8.195. Other key discoveries include $ponA1$ (0.308) and $pknB$ (0.273) for RIF, $mmpL5$ (4.947) for RFB, and the novel $lipG$ (0.756) and $idsA2$ (0.536) loci for EMB. The consistent prioritization of these novel loci across all four drugs suggests that VAMP-Net’s attention mechanism successfully isolates rare, high-effect mutations and epistatic modifiers that are often missed by standard frequency-based methods.

\subsubsection{Functional Enrichment}

To determine whether the high-attribution features identified by VAMP-Net reflect genuine biological mechanisms, we cross-referenced the top-ranked variants against established proteomic and molecular literature. As detailed in Table~\ref{tab:biological_validation}, our framework successfully identified a hierarchical resistance signature that spans from canonical target modification to global cell-envelope remodeling. VAMP-Net correctly prioritized \textit{rpoB} variants, specifically distinguishing the unique structural binding affinities of Rifabutin (RFB) over Rifampicin (RIF) through the selection of variants like 761140\_A$>$T (IG: 6.896) \citep{jamieson2014profiling}. Beyond canonical targets, the framework isolated broad-spectrum efflux mechanisms, notably the \textit{mmpL5} system (IG: 4.947), which likely modulates RFB susceptibility through active drug extrusion \citep{hartkoorn2014cross}. Additionally, the high ranking of the persistence-related \textit{lpqS} gene (IG: 4.213) suggests the model captures the evolutionary trajectory of resistance where enhanced macrophage survival precedes the accumulation of primary resistance markers \citep{sakthi2016lipoprotein} Significantly, the model identified an integrated metabolic program for cell-wall reinforcement. High-attribution scores for \textit{ppsD}, \textit{fadD26}, and \textit{papA4}, members of a co-transcribed hub, reflect the biosynthesis of the Phthiocerol Dimycocerosates (PDIM) waxy coat \citep{bisson2012upregulation, simeone2010delineation}. The prioritization of these loci, known to be upregulated in \textit{rpoB} mutants, demonstrates that VAMP-Net has internalized the compensatory permeability barriers employed by \textit{M. tuberculosis} under drug pressure. These results confirm that the framework reconstructs a biologically grounded, multi-layered defense strategy rather than relying on arbitrary statistical correlations.  \\

\begin{table*}[htbp]
\centering
\caption{Predictive attribution and biological functional validation of representative Rifabutin (RFB) resistance markers identified by VAMP-Net.}
\label{tab:biological_validation}
\small
\begin{tabularx}{\textwidth}{@{} llcl X r @{}}
\toprule
\textbf{Gene} & \textbf{Variant} & \textbf{IG Score} & \textbf{Category} & \textbf{Functional Mechanism} & \textbf{Reference} \\ \midrule

\textit{rpoB} & 761140\_A$>$T & 6.896 & Primary & Canonical RNA Polymerase target; different structural & \citep{jamieson2014profiling} \\

 & 761139\_C$>$A & 3.955 & Resistance & affinities for RFB vs. RIF yield distinct mutation profiles. & \\ \addlinespace

\textit{mmpL5} & 2944085\_C$>$T & 4.947 & Efflux Pump & Broad-spectrum efflux (Bedaquiline/Clofazimine); & \citep{hartkoorn2014cross} \\
 & & & & likely contributes to non-specific RFB resistance. &  \\ \addlinespace

\textit{lpqS} & 893702\_GAC... & 4.213 & Persistence & Resists copper toxicity in macrophages; enhances & \citep{sakthi2016lipoprotein} \\
 & & & & persistence and subsequent mutation accumulation. & \\ \addlinespace

\textit{ppsD} & 3640407\_G$>$A & 3.532 & Cell Wall & Essential for PDIM (waxy coat) biosynthesis; directly & \citep{bisson2012upregulation} \\
 & & & (PDIM) & upregulated in \textit{rpoB} mutants to enhance permeability barrier. & \\ \addlinespace

\textit{fadD26} & 3545402\_A$>$G & 3.527 & Cell Wall & Initiates PDIM biosynthesis; co-transcribed with \textit{ppsD} & \citep{simeone2010delineation} \\
 & & & (PDIM) & to reinforce the outer waxy envelope. & \\ \addlinespace

\textit{pks13} & 1262023\_C$>$T & 2.885 & Cell Wall & Final catalyst for mycolic acid synthesis; primary & \citep{gavalda2009pks13} \\
& 1262193\_C$>$T & 2.589 & (Mycolic) & barrier layer essential for natural drug resistance. & \\ \addlinespace

\textit{papA4} & 4217696\_TGG.. & 2.594 & Cell Wall & Anchors PDIM to the surface; loss of function leads & \citep{hood2001identification} \\
 & & & (PDIM) & to envelope mislocalization and drug vulnerability. & \\
\bottomrule
\end{tabularx}
\end{table*}

\begin{figure}[htbp]
\centering
\includegraphics[width=0.95\textwidth,height=0.78\textheight,keepaspectratio]{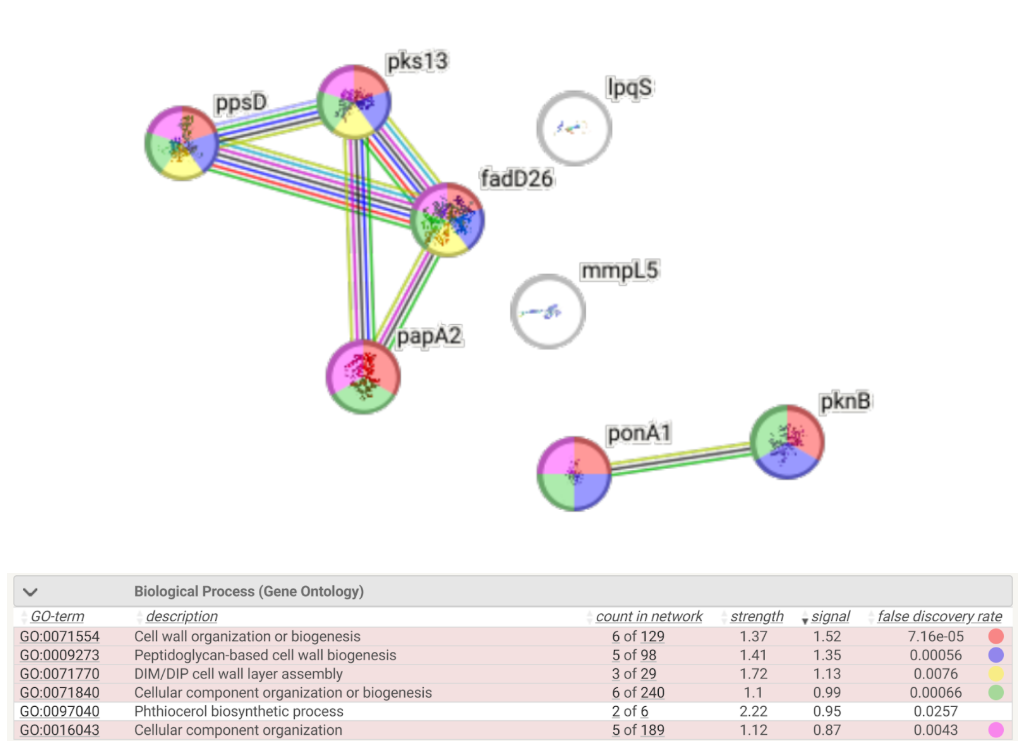}
\caption{Functional protein-protein interaction network and GO enrichment of novel loci identified for RIF/RFB resistance. The interaction map (top) highlights two significant metabolic modules discovered via VAMP-Net`s attention weights: the ppsD-pks13-fadD26 cluster and the ponA1-pknB axis. The Gene Ontology analysis (bottom) confirms these loci are statistically enriched for cell wall organization ($p = 7.16 \times 10^{-5}$) and PDIM biosynthesis ($p = 0.0257$). These results demonstrate VAMP-Net's ability to recover biologically cohesive compensatory mechanisms alongside canonical resistance markers.
}
\label{fig:rfb_enr}
\end{figure}

To validate the biological coherence of the identified novel resistance loci, we performed functional enrichment analysis using STRING-db \ref{fig:rfb_enr}. The high-attribution genes (\textit{ponA1, pknB, mmpL5, lpqS, ppsD, fadD26, pks13, papA2}) formed a statistically significant interaction network ($p = 0.00239$), confirming they constitute a biologically non-random module. Gene Ontology (GO) analysis identified six enriched biological processes converging on cell wall remodeling as a unified adaptive strategy: Significant enrichment in "Cell wall organization or biogenesis" ($FDR = 7.16 \times 10^{-5}$) and "Peptidoglycan-based cell wall biogenesis" ($FDR = 0.00056$) directly validates the RIF-associated loci \textit{ponA1} and \textit{pknB}; enrichment in "DIM/DIP cell wall layer assembly" ($FDR = 0.0076$) and "Phthiocerol biosynthetic process" ($FDR = 0.0257$) confirms that the model autonomously identified the precise PDIM/phthiocerol lipid pathways. High enrichment strength for the "Phthiocerol biosynthetic process" (strength = 2.22) biochemically confirms the functional assembly of \textit{ppsD}, \textit{fadD26}, and \textit{pks13}.These results demonstrate that VAMP-Net`s discovery reflects the coordinated upregulation of PDIM biosynthesis and cell wall reinforcement as an adaptive epistatic response to \textit{rpoB}-mediated resistance.\\

\begin{figure}[htbp]
\centering
\includegraphics[width=0.95\textwidth]{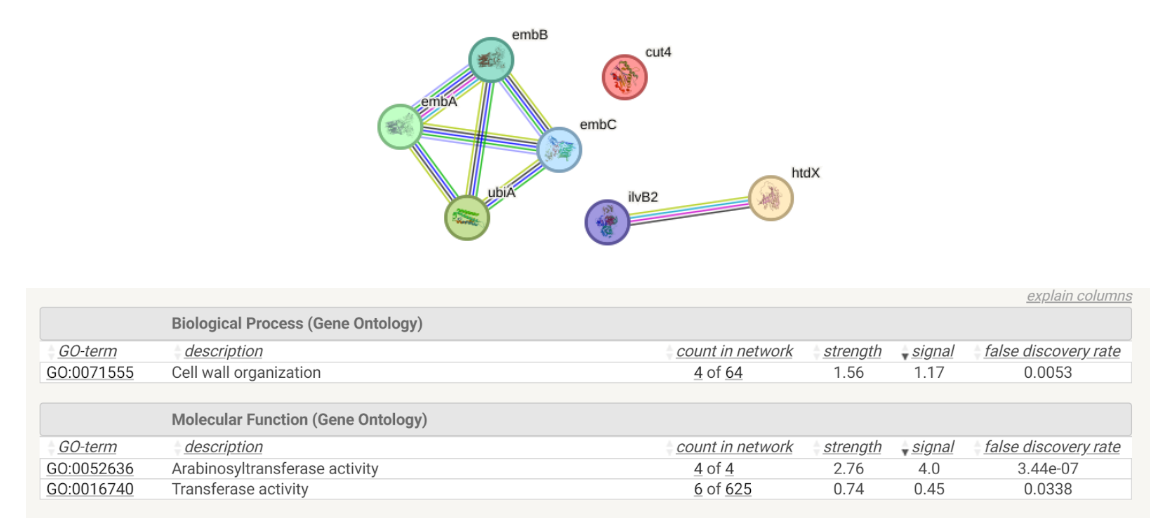}
\caption{Functional interaction network and GO enrichment for Ethambutol (EMB) resistance loci. The interaction map (top) illustrates a tightly coupled module comprising \textit{embA, embB, embC,} and \textit{ubiA}, which are primary drivers of EMB resistance. Gene Ontology analysis (bottom) confirms high statistical enrichment for arabinosyltransferase activity ($p = 3.44 \times 10^{-7}$) and cell wall organization ($p = 0.0053$). These results validate VAMP-Net's ability to precisely identify the specific metabolic pathways targeted by the drug.
}
\label{fig:emb_enr}
\end{figure}

Functional enrichment analysis of EMB-associated loci yielded the study`s strongest validation result \ref{fig:emb_enr}, centered on Arabinosyltransferase activity (GO:0052636). This term achieved an exceptional enrichment signal of 4.0 ($FDR = 3.44 \times 10^{-7}$), capturing 4 out of 4 known arabinosyltransferase-encoding genes in the M. tuberculosis genome.This constitutes a near-perfect biological validation, as ethambutol (EMB) specifically targets the arabinosyltransferase enzymes, encoded by embA, embB, and embC—responsible for cell wall synthesis. VAMP-Net autonomously recovered this entire enzyme family along with the ubiA locus, a known resistance gene providing the necessary substrate. Furthermore, "Cell wall organization" (GO:0071555) was significantly enriched ($FDR = 0.0053$), consistent with EMB's role in disrupting cell wall assembly. These findings confirm that the model learns integrated resistance pathways, comprising primary targets, substrate supply, and potential epistatic modifiers, rather than isolated variant associations.

\subsubsection{Epistatic Interaction Network Analysis}
\label{subsubsec:epistasis_analysis}

\begin{figure}[htbp]
\centering
\includegraphics[width=0.65\textwidth]{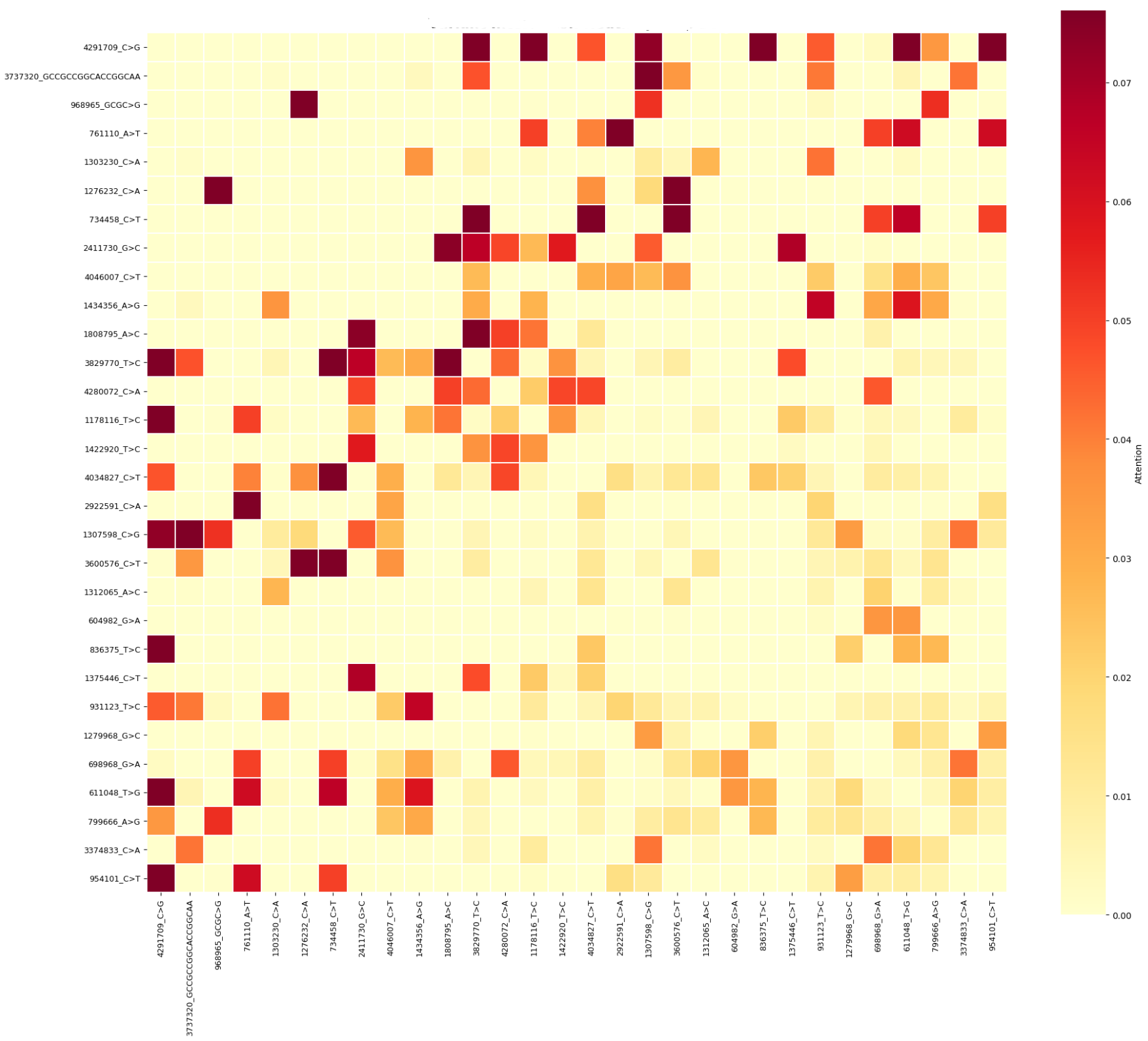}
\caption{Interpretable attention heatmap of variant-variant interactions for Rifampicin (RIF) resistance. The matrix visualizes the self-attention weights derived from the Set Attention Block (SAB), representing the perceived strength of interaction between genomic variants. Darker red clusters identify "hub variants" that exert high influence over the collective resistance profile. These patterns reveal that VAMP-Net identifies non-linear epistatic relationships beyond simple additive effects, aligning with the biological understanding that complex resistance phenotypes often emerge from the co-occurrence of specific compensatory and primary mutations.}
\label{fig:variant_heatmaps_rif}
\end{figure}

Analysis of the aggregated Set Attention Block (SAB) weights culminates in the Multi-locus Attribution-Weighted Interaction (MAWI) Map.
This approach quantifies the synergistic coordination between mutations, moving beyond single-variant associations to decompose the polygenic architecture of resistance. The interaction heatmap for Rifampicin (RIF) reveals a highly interconnected network with high signal density (Figure \ref{fig:variant_heatmaps_rif}). 
This structure is characterized by "hub" variants, such as 3829770\_T$>$C and 1307598\_C$>$G, which exhibit strong cross-attention with multiple secondary mutations. This hub-driven topology quantitatively supports the hypothesis of epistatic fitness compensation, where robust resistance relies on a coordination of compensatory pathways to mitigate the fitness costs of primary mutations.

In contrast, the Rifabutin (RFB) network 
exhibits a sparse, localized architecture. The attention landscape is dominated by a few high-leverage interactions involving variants like 1070702\_T$>$C, set against a low-interaction background. This sparsity indicates that RFB resistance is governed by a select few non-additive synergistic effects that are mechanistically determinative, rather than the diffuse polygenic mechanism observed in RIF.


The contrasting patterns between these two rifamycins suggest that while they share target genes, their resistance modules are distinct. VAMP-Net’s ability to pinpoint these drug-specific interaction hubs provides an interpretable, auditable trace of the model’s reasoning. These results confirm known biological dependencies while highlighting novel epistatic interaction points that could serve as targets for future functional validation and the design of more effective combination therapies.

To confirm that the high-attention mutation pairs identified by the SAB transformer reflect genuine genomic dependencies rather than modeling artifacts, we performed a population-level validation using Fisher’s exact test and Odds Ratio (OR) analysis (Table \ref{tab:epistasis_stats}). This approach provides an external statistical benchmark for the non-random dependencies captured by the attention mechanism. As detailed in Table Table \ref{tab:epistasis_stats}, the statistical results for the primary interaction candidates demonstrate robust data-driven support:

\begin{itemize}
    \item RIF-associated pair ($1276232:C{>}A \leftrightarrow 3600576:C{>}T$): The combination of an infinite OR and an extremely low $p$-value ($9.296 \times 10^{-89}$) indicates a highly structured co-occurrence. This suggests possible genetic linkage or strong co-selection under drug pressure, where the presence of one mutation is effectively determinative of the other.
    
    \item RFB-associated pair ($1070702:T{>}C \leftrightarrow 985287:G{>}A$): The elevated odds ratio of $8.18$ ($p < 0.01$) indicates significant enrichment, confirming that these mutations co-occur substantially more frequently than expected under independence.
\end{itemize}

These findings provide a critical layer of interpretability for VAMP-Net. By demonstrating that the highest-weighted interaction hubs are backed by significant population-level enrichment, we confirm that the model recovers mutation pairs consistent with the underlying structure of the \textit{M. tuberculosis} population. This validates the attention heatmap as an auditable biological trace, effectively pinpointing non-random dependencies that may serve as targets for further functional investigation into epistatic fitness or shared resistance backgrounds.

\begin{table}[htbp]
\centering
\caption{Statistical validation of representative high-attention mutation pairs using Fisher's exact test. Only mutation pairs with strong statistical support are shown.}
\label{tab:epistasis_stats}
\footnotesize
\setlength{\tabcolsep}{4pt}
\renewcommand{\arraystretch}{1.12}
\begin{tabularx}{\textwidth}{l>{\raggedright\arraybackslash}p{4.1cm}cc>{\raggedright\arraybackslash}X}
\toprule
\textbf{Drug} & \textbf{Mutation pair} & \textbf{OR} & \textbf{$p$-value} & \textbf{Interpretation} \\
\midrule
RIF & $1276232:{C{>}A} \leftrightarrow 3600576:{C{>}T}$ & $\infty$ & $9.296 \times 10^{-89}$ & Extremely strong non-random co-occurrence; consistent with tight genetic linkage, co-selection, or a shared resistance background. \\

RFB & $1070702:{T{>}C} \leftrightarrow 985287:{G{>}A}$ & 8.18 & $7.175 \times 10^{-3}$ & Significant enrichment, indicating a robust association beyond chance. \\
\bottomrule
\end{tabularx}
\end{table}


\subsubsection{VCF Feature Channel Analysis}

\begin{figure}[htbp]
\centering
\includegraphics[width=1\textwidth]{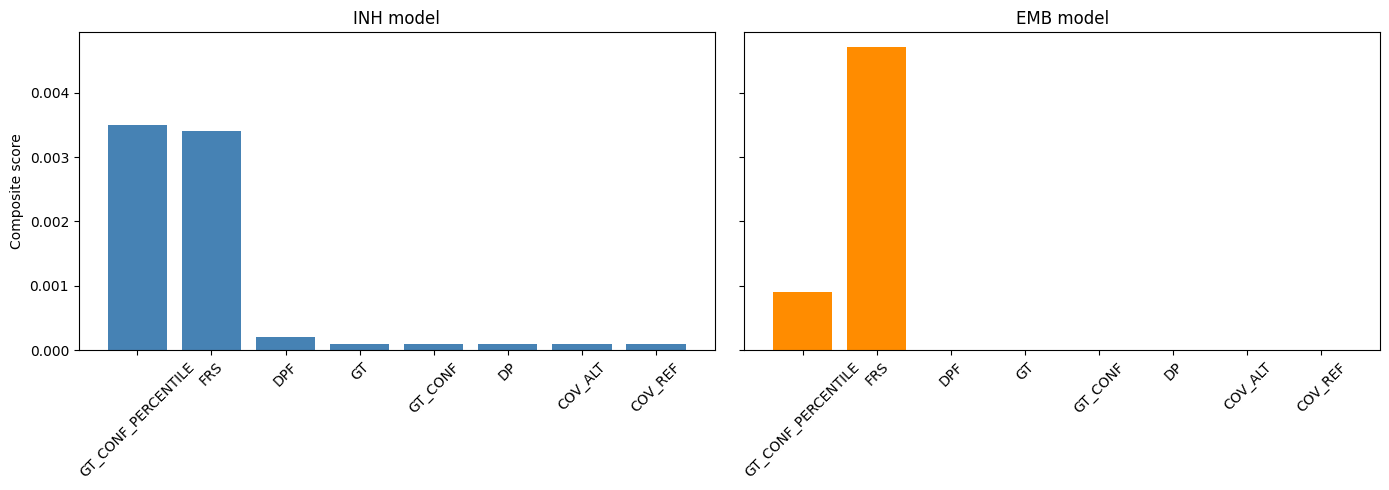}\\[0.5em]
\includegraphics[width=1\textwidth]{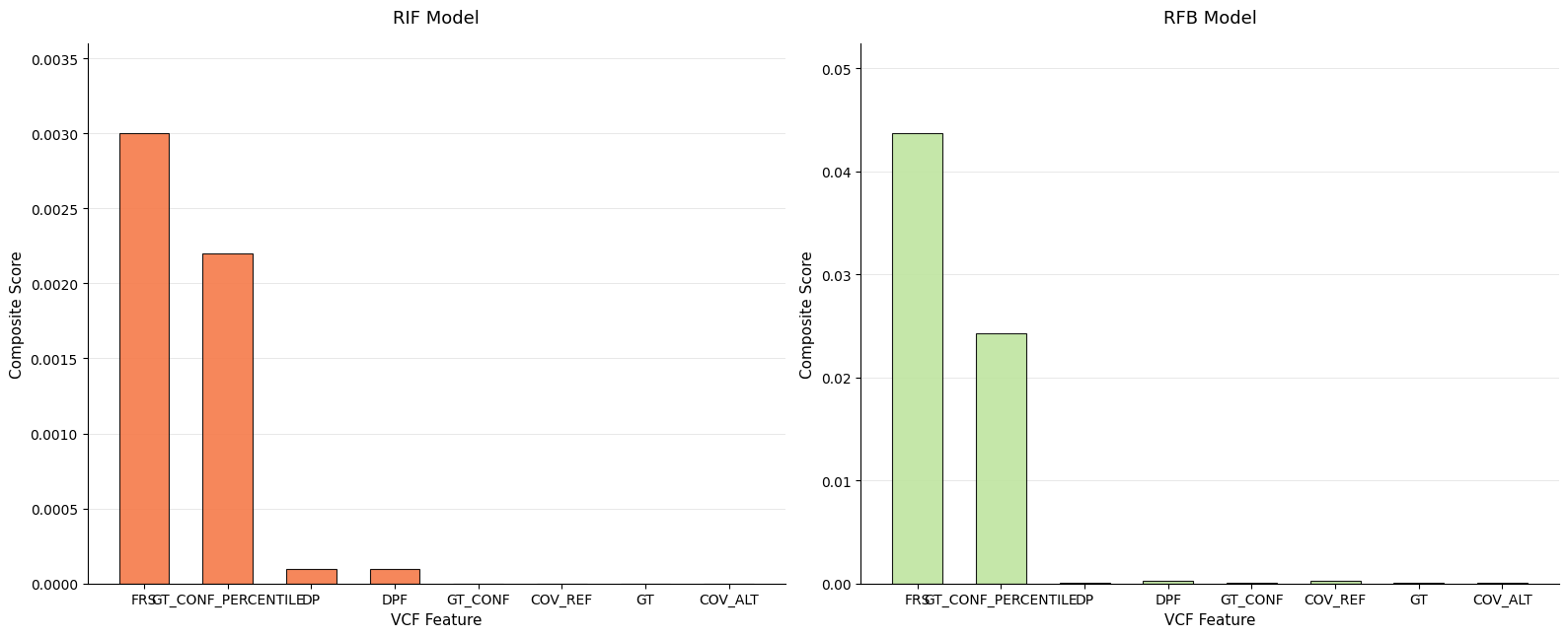}
\caption{Quantitative importance of technical quality metrics across EMB, INH, RIF, and RFB resistance prediction. The upper panel compares EMB (orange) and INH (blue), while the lower panel compares RIF (orange) and RFB (green). Across phenotypes, Fraction of Supporting Reads (FRS) and Genotype Confidence Percentile (GT\_CONF\_PERCENTILE) emerge as dominant technical predictors, indicating that VAMP-Net consistently leverages sequencing-confidence cues to recalibrate resistance predictions.}
\label{fig:vcf_features}
\end{figure}

To validate the multi-channel design of the Quality-Aware 1D-CNN pathway, we quantified the contribution of eight VCF features through systematic ablation (Figure \ref{fig:vcf_features}). The analysis confirms that the model transcends simple variant presence (\textit{GT}) by leveraging granular contextual metrics:

\begin{itemize}
    \item Dominant Metric (\textit{FRS}): The Fraction of Supporting Reads (\textit{FRS}) is the most influential feature across both RIF and RFB models. This confirms the pathway effectively filters low-support variants, aligning with clinical standards for calling clonal mutations.

\item Contextual Confidence: \textit{GT\_CONF\_PERCENTILE} serves as the second most critical feature. By prioritizing relative confidence ranks over raw \textit{GT\_CONF} values, the model demonstrates a sophisticated capacity to utilize sample-specific comparative context.

\item Redundancy of Raw Metrics: The near-zero scores for Read\_Depth (\textit{DP}) and Genotype (\textit{GT}) suggest that their information is successfully subsumed by normalized, derived features.
\end{itemize}

These results demonstrate that the 1D-CNN pathway successfully extracts technically meaningful information, enabling a robust distinction between true resistance-conferring variants and sequencing noise.

\subsection{Discussion and Limitations}
\label{subsec:discussion}

\subsubsection{Architectural Efficacy and Integration Rationale}

VAMP-Net demonstrates exceptional predictive fidelity across the anti-TB drug cohort, specifically achieving peak ROC-AUCs of 0.975 (RFB) and 0.970 (INH). The remarkably low variance across experimental seeds (std $<$ 0.02) confirms that this performance is a robust consequence of the model's structural design rather than stochastic initialization. Even in the presence of significant class imbalance, such as with Ethambutol (19.8\% resistant), the model achieved the cohort’s highest PR-AUC (0.977), proving its resilience to the "curse of dimensionality" and skewed genomic datasets common in clinical settings. Comparative benchmarking reveals a significant performance gap between VAMP-Net and traditional baselines, with the MLP failing to exceed 85.88\% accuracy. While the CNN effectively captured local spatial patterns, its higher false-negative rate (10.6\%) relative to VAMP-Net (7.1\%) underscores the necessity of attention mechanisms for modeling long-range epistatic interactions. VAMP-Net’s superior sensitivity in identifying 92.9\% of resistant isolates while maintaining high specificity (91.3\%) validates its utility as a high-precision diagnostic tool for complex antimicrobial resistance landscapes. Ultimately, the model’s success is rooted in its late-stage adaptive fusion philosophy. In contrast to Early Fusion, which suffered from a catastrophic specificity deficit (0.425) due to premature noise propagation, VAMP-Net’s multi-path design allows for modality-specific representational learning. By postponing integration until after contextual encoding, the framework achieved transformative gains, including a +48.8\% increase in Specificity and a +30.9\% increase in ROC-AUC. These results provide a mathematically grounded justification for VAMP-Net as a stable, noise-resilient framework for rapid molecular-based tuberculosis diagnostics.


\subsubsection{VCF-Aware Recalibration and Clinical Utility}

The granular case studies presented in Table 11 provide a critical window into the decision-making process of VAMP-Net, highlighting the Path-2 branch as a dynamic reliability filter. The substantial magnitude of observed probability shifts ($\Delta P \approx$ - 0.86) indicates that VCF-level metadata does not merely refine the genomic signal but fundamentally re-weights evidence based on technical fidelity. In Corrected cases, while Path-1 identifies resistance-like mutation patterns, Path-2 detects a conflict where weak sequencing support, characterized by low \textit{GT\_CONF\_PERCENTILE} or insufficient read depth, renders the initial call unreliable. Conversely, in Amplified cases, Path-2 acts as a force-multiplier for high-quality data, shifting borderline predictions ($P(R)\approx$ 0.55) to high-confidence resistant calls ($>$ 0.90). By distinguishing between true biological polymorphisms and sequencing artifacts, VAMP-Net effectively treats variant presence as a necessary, but not sufficient, condition for resistance prediction. Beyond raw accuracy, VAMP-Net addresses the practical barriers to AI adoption in healthcare: calibration and operational efficiency. Brier score analysis confirms that VAMP-Net is a well-calibrated classifier, particularly for INH ($\sim 0.05$). High calibration ensures that predicted probabilities reflect the true likelihood of resistance, thereby reducing the clinical risk of suboptimal patient treatment. Furthermore, the architecture maintains a high-speed inference rate of $\sim490$ samples per second. This computational efficiency, combined with high reproducibility and the reduction of \textit{grey-zone} predictions through Path-2 recalibration, positions VAMP-Net as a reliable foundational tool for the next generation of rapid, molecular-based tuberculosis diagnostics.

\subsubsection{Interpretability and Mechanistic Discovery}

VAMP-Net’s dual-path architecture provides a mathematically grounded framework for identifying high-impact resistance mechanisms by effectively bridging high-dimensional genomics with biological and technical insights. Feature attribution analysis successfully recovered canonical resistance signatures, such as rpoB variants for RIF and RFB (e.g., 761140\_A$>$T; Score: 6.896), while internalizing complex co-selection patterns in MDR-TB strains by prioritizing cross-drug markers like katG (Score: 0.446). Beyond known targets, the Set Attention Block ($SAB$) identified novel, high-effect candidates—including the mmuM locus for INH (Score: 8.195) and the mmpL5/ppsD axis for RFB—which functional enrichment confirms constitute a statistically significant ($p = 0.00239$) metabolic module centered on cell-wall remodeling and PDIM biosynthesis. Furthermore, systematic ablation of VCF features validates the 1D-CNN's role as a signal-to-noise regulator that prioritizes the Fraction of Supporting Reads (\textit{FRS}) and relative confidence rankings over raw depth. This "integrated audit" ensures that final predictions are governed by a synergy of genomic pathogenicity and technical reliability, allowing the model to dynamically down-weight biologically relevant variants that lack sufficient sequencing support. Collectively, these findings position VAMP-Net as an auditable discovery tool capable of reconstructing the multi-layered defense strategy of M. tuberculosis through the lens of both evolutionary selection and clinical data quality.


\subsubsection{Limitations of the Study}

While VAMP-Net demonstrates high predictive and interpretability performance, its scope is bounded by several factors. First, although our tokenization strategy effectively captures rare variants, generalizability remains constrained by the training cohort's diversity; emerging mutations or rare lineages absent from the dataset may degrade performance. Second, while the 1D-CNN pathway successfully filters technical noise, it does not currently address broader systemic errors such as metagenomic contamination or sample-swapping in clinical workflows. Finally, the attention-based feature attribution provides a computational "causality hypothesis" rather than biochemical proof. The identified epistatic interactions and novel loci, though statistically significant, require experimental validation through site-directed mutagenesis or functional assays to confirm their mechanistic roles


\section{Conclusion}
\label{conclusion}

This study introduced the Interpretable Variant-Aware Multi-Path Network (VAMP-Net), a model designed to bridge the gap between high-dimensional genomic pathogenicity and technical data reliability. By integrating a Permuutation-Invariant Set Attention Transformer for epistatic modeling with a Quality-Aware 1D-CNN, VAMP-Net addresses the critical challenge of separating biological signals from sequencing noise. Our results demonstrate that the dual-path architecture successfully recovers canonical resistance markers while discovering novel loci, such as mmuM and mmpL5, within statistically significant metabolic modules centered on cell-wall remodeling. The most compelling methodological advancement is the model's capacity for adaptive feature calibration. Quantitative attribution analysis confirms that the fusion mechanism dynamically modulates its reliance on quality metrics (e.g., FRS and GT\_CONF\_PERCENTILE) based on the phenotypic complexity, providing an unprecedented level of decision robustness for subtle resistance profiles like Rifabutin. This "integrated audit" ensures that all predictions are supported by both biological and technical evidence streams, establishing a new standard for transparency in clinical AI.\\

Future work will focus on extending this multi-path paradigm to multi-modal clinical settings. By incorporating longitudinal metadata and medical imaging (e.g., histology or microscopy) alongside genomic sets, we aim to evolve VAMP-Net into a comprehensive diagnostic engine. Ultimately, this work provides a scalable, mathematically grounded template for genotype-phenotype association in any domain characterized by variable-length inputs, complex epistasis, and heterogeneous data quality.


\section*{Acknowledgement}

This work was carried out partially using the AI Datacenter of the National School of Artificial Intelligence, funded under grant number E049 24 0117 by the Algerian Ministry of Higher Education and Scientific Research.

\section*{Data availability statements}
The complete dataset used in this study is publicly available at \url{ftp.ebi.ac.uk/pub/databases/cryptic/release_june2022/}.

\section*{Code availability }
The complete code of this study can be  found at \url{https://github.com/Aicha-AI/VAMP-Net.MTB-Drug-Resistance/tree/main}.



\bibliographystyle{plainnat} 
\bibliography{references}

\end{document}